%% file: main.tex
\documentclass{article}

\usepackage{iclr2026_conference,times}
\usepackage{latexsym}
\usepackage[T1]{fontenc}
\usepackage[utf8]{inputenc}
\usepackage{microtype}
\usepackage{inconsolata}
\usepackage{graphicx}
\usepackage{subcaption}
\usepackage{float}
\usepackage{placeins}
\usepackage{wrapfig}
\usepackage{booktabs}
\usepackage{multirow}
\usepackage{amsmath}
\usepackage{amssymb}
\usepackage{xcolor}
\usepackage{hyperref}
\usepackage{url}
\usepackage{listings}
\usepackage[most]{tcolorbox}
\tcbuselibrary{listings,breakable,skins}
\lstdefinestyle{promptstyle}{
  basicstyle=\ttfamily\scriptsize,
  breaklines=true,
  breakatwhitespace=false,
  columns=flexible,
  showstringspaces=false,
  breakindent=0pt
}
\newtcblisting{promptboxinline}[2][]{%
  enhanced, breakable, listing only,
  listing engine=listings,
  listing options={style=promptstyle},
  colback=gray!5, colframe=gray!40, coltitle=black,
  colbacktitle=gray!14, fonttitle=\bfseries\footnotesize,
  boxrule=0.4pt, arc=2pt, left=1.2mm, right=1mm, top=1mm, bottom=1mm,
  before skip=0.9em, after skip=1.1em,
  title={#2},
  title after break={#2 \normalfont(continued)},
  #1
}

\newtcolorbox{placeholderbox}[2][]{%
  colback=black!4, colframe=black!25, boxrule=0.4pt, arc=2pt,
  halign=center, valign=center, height=#2, #1}

\graphicspath{{figures/}}

\title{FigMirror: Ground It, Code It, Plot It}

\author{
  Xiaohan Zhao\textsuperscript{*}
  \quad Jiacheng Liu\textsuperscript{*}
  \quad Yaxin Luo
  \quad Zhiqiang Shen\textsuperscript{\textdagger} \\
  \normalfont VILA Lab, Department of Machine Learning, MBZUAI \\
  \normalfont\small
  \textsuperscript{*}Equal contribution.\qquad
  \textsuperscript{\textdagger}Corresponding author:
  \texttt{zhiqiang.shen@\allowbreak mbzuai.ac.ae}
}

\hypersetup{
  pdftitle={FigMirror: Ground It, Code It, Plot It},
  pdfauthor={Xiaohan Zhao, Jiacheng Liu, Yaxin Luo, Zhiqiang Shen}
}

\iclrfinalcopy

\begin{document}
\maketitle
\lhead{Preprint}

\begin{abstract}
Converting scientific figures into executable code has gained increasing attention, yet existing methods primarily focus on reproducing the reference figure itself. A more practical setting is to plot new data while preserving the visual style of a reference figure (e.g., color scheme and typography). Prior approaches mimic the reference through pixel-level optimization and struggle to carry its style to new data. We show that the key to this task lies in the coordinate grounding and coding capabilities present in modern computer-use models.
We propose \texttt{FigMirror}, an agentic framework that unlocks these capabilities through {\em Grounded Measurement}, which locates visual elements by coordinates and measures their properties through executable code. We further introduce \texttt{PlotTwin-Bench}, an expert-curated benchmark with fine-grained code and image-level style metrics. Experiments show that \texttt{FigMirror} consistently outperforms existing methods on reference-conditioned style transfer. All plots in this paper are generated by \texttt{FigMirror}, except those produced by other methods for comparison. Our code and data are available at: \url{https://github.com/VILA-Lab/FigMirror}.
\end{abstract}

\begin{figure}[H]
  \centering
  \begin{subfigure}[t]{0.4600\linewidth}
    \centering
    \includegraphics[width=\linewidth]{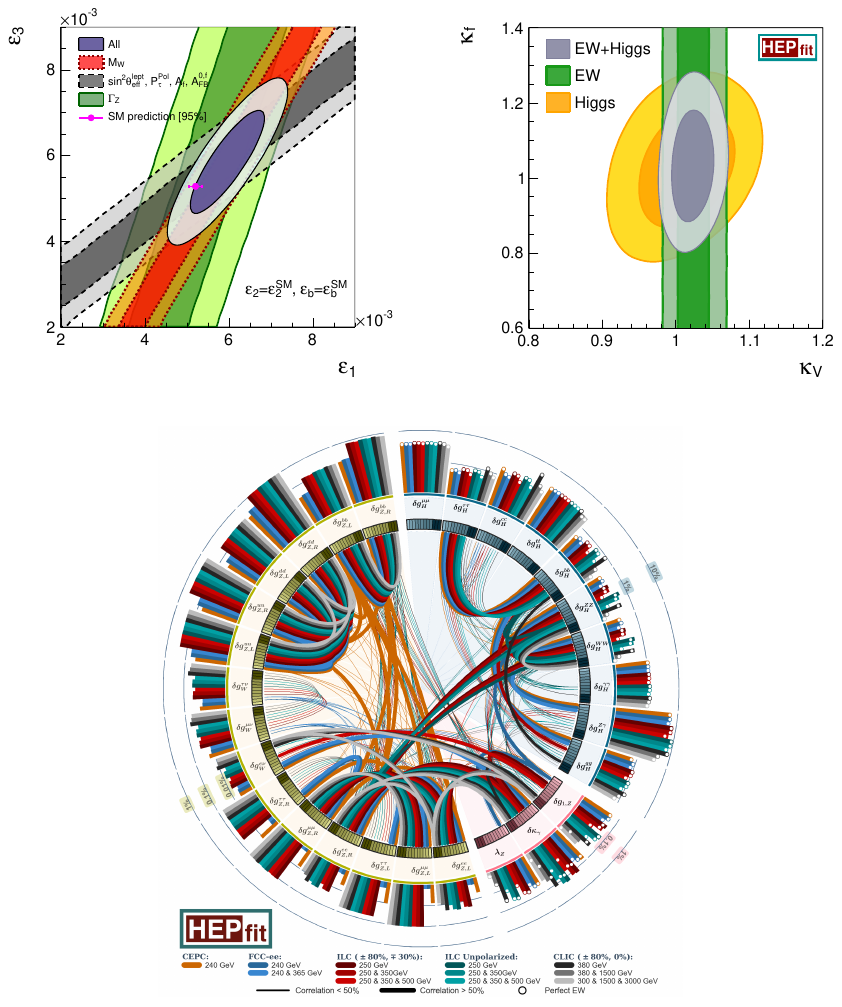}
    \caption{Reference~\citep[Fig.~1]{de2020hepfit}}
    \label{fig:teaser-reference}
  \end{subfigure}%
  \begin{subfigure}[t]{0.4308\linewidth}
    \centering
    \includegraphics[width=\linewidth]{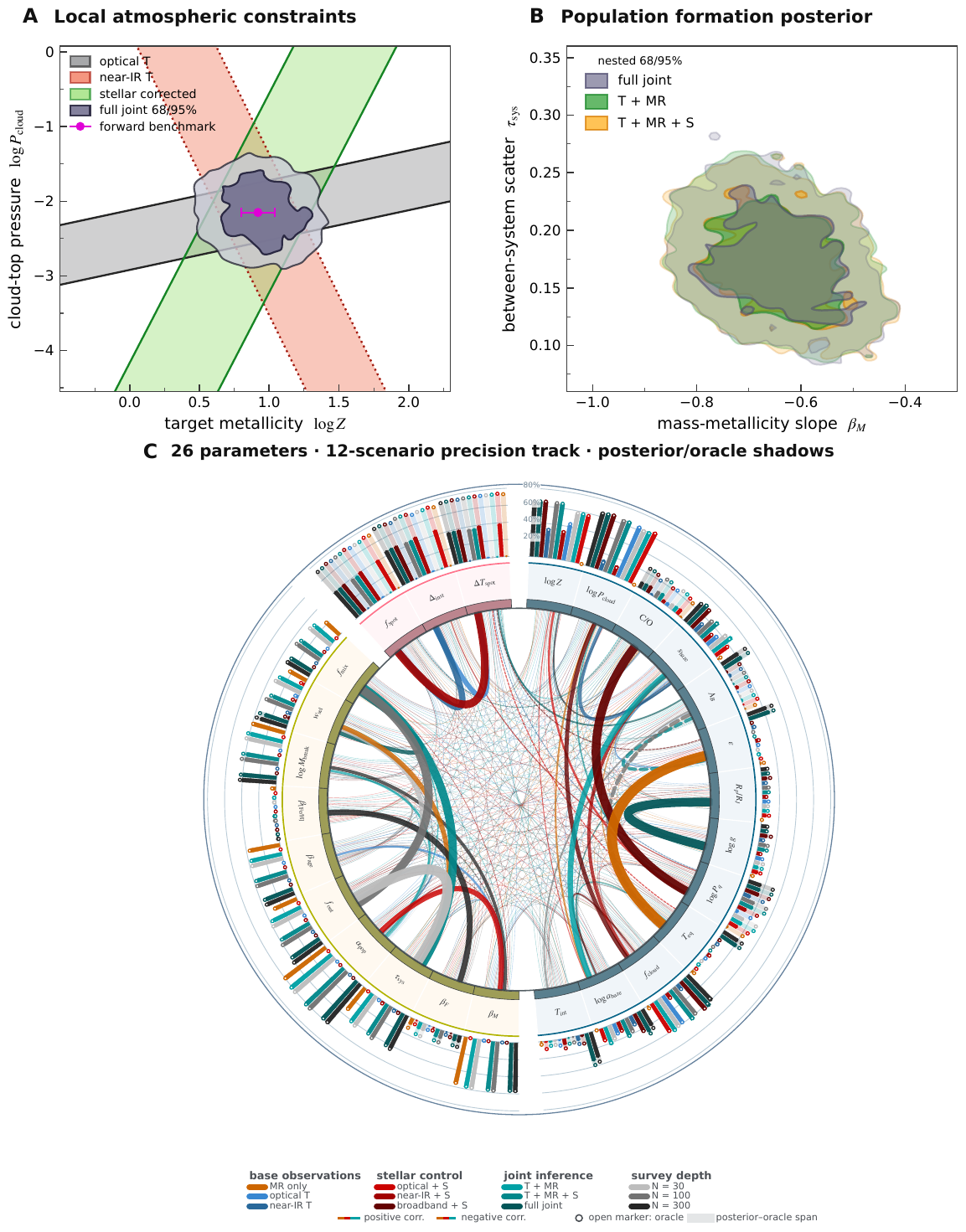}
    \caption{Style transfer by FigMirror}
    \label{fig:teaser-output}
  \end{subfigure}
  \caption{\textbf{Reference-conditioned style transfer.} \texttt{FigMirror} preserves the visual style of a reference scientific figure while
  adapting it to new target data.}
  \label{fig:teaser}
\end{figure}

\input{sections/01_introduction}
\input{sections/02_related_work}
\input{sections/03_method}

\input{sections/04_experiments}

\FloatBarrier
\section{Conclusion}

We study reference-conditioned style transfer for scientific figures.
Because the target data differ from the reference, transferable style
cannot be assessed through whole-image matching; its elements must
instead be localized, measured, and reapplied. Our main
insight is that scientific figures share the code-rendered structure of
graphical interfaces: both are built from discrete elements with sharp
boundaries and flat fills. The coordinate grounding that lets a
model point to a button or a menu entry therefore transfers
to plot elements such as ticks, legends, and marks.

We turn this insight into Grounded Measurement, which locates each style
element and reads its value with code, and we organize generation as a
Drawer--Reviewer loop that makes every style attribute explicit,
measurable, and revisable. To evaluate this setting, we build \texttt{PlotTwin-Bench}, the first
benchmark for scientific figure style transfer, and score each candidate
from both the code and the rendered image against per-reference
scoring criteria. We hope this formulation makes figure style a measurable and
reusable design choice.

\section*{Ethics Statement}
\input{sections/07_ethics}

\clearpage
\appendix
\input{appendix/appendix}

\end{document}

%% file: sections/01_introduction.tex
\section{Introduction}

Producing a publication-quality scientific figure is laborious and challenging. The spacing, panel sizes, and fonts often require multiple rounds of manual adjustment before the figure looks right~\citep{rougier2014ten}. Researchers have traditionally used well-crafted figures from top venues as references and manually adopted their spacing, colors, and layouts. Multimodal models can now interpret such references and generate the corresponding plotting code. However, most existing chart-to-code methods focus on reproducing the reference figure itself at the pixel level, rather than transferring its visual style and layout to new data.

Chart-to-code generation predates capable general-purpose large models, and early work fine-tuned specialized models for this task. Reproduction provides a natural training signal: the discrepancy between the rendered figure and the reference. ChartLlama~\citep{han2023chartllama} is trained in this manner. General-purpose models have since made substantial progress in visual understanding and reasoning, yet the same recipe persists. METAL~\citep{li2025metal} adopts a multi-agent framework in which a generator writes the code and a critic repeatedly compares the rendered output against the reference and revises the code accordingly. This effectively moves the same optimization process to inference time. In either case, the objective remains reproduction. By contrast, transferring visual style of a reference figure to a researcher's own data has received little attention as a task in its own right.



\begin{figure}[t]
  \centering
  \includegraphics[width=0.95\linewidth,trim={0pt 8pt 0pt 8pt},clip]{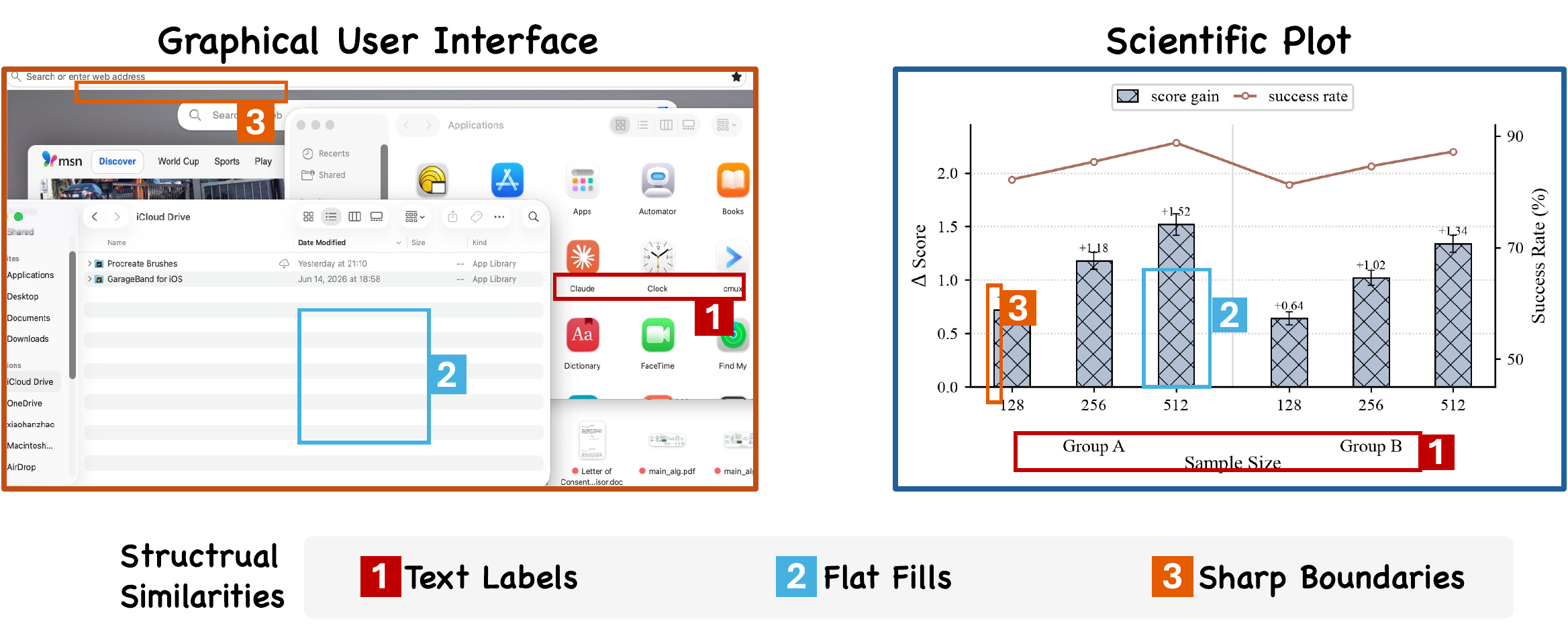}
  \vspace{-0.2cm}
  \caption{GUIs and scientific figures share localized, code-rendered
  elements with sharp boundaries, flat fills, and text. This structural
  similarity allows coordinate grounding learned for GUI elements to
  localize plot elements such as ticks, legends, and marks for
  measurement and review.}
  \label{fig:gui-plot-comparison}
\end{figure}

A figure entangles its data with its style. We argue that transferring the style requires disentangling the two in two steps. The first is \textit{identification}: locating the visual attributes that carry the transferable style, e.g., a color series or the spacing between subplots. The second, more critical, is \textit{measurement}: computing each attribute's exact value, e.g., a hex code or a width in points, from the element's pixels. Modern general-purpose models have already learned the first step, just not from plotting. They are trained at scale for computer use~\citep{anthropic2026opus5,openai2026gpt56,bai2025qwen3,wang2025ui}. There, an agent operates software on behalf of a user, and every action begins with the coordinates of its on-screen target, a button or a menu entry. This demand instills coordinate grounding. A scientific plot, built from discrete, sharp-edged elements rendered by code (Figure~\ref{fig:gui-plot-comparison}), is as amenable to coordinate grounding, and as readily as a GUI. 
Yet existing plotting tasks do not explicitly invoke this skill or capability. When asked to match a reference figure, the model still tends to estimate visual attributes by eye.

To address this, we propose {\em Grounded Measurement} to redirect the skill from acting to measuring. The model grounds the element that carries a style attribute, and where a click would follow, a short program reads the exact value, a line's width or a series' color. One measurement, however, resolves one attribute. A transfer needs every attribute found, measured, applied, and re-checked on the rendered figure. Our proposed framework, \texttt{FigMirror}, organizes this work as a {\bf Drawer--Reviewer} loop. The Drawer decomposes the reference into style attributes, measures each, and renders the user data into a candidate. The Reviewer checks the candidate against the reference visually. The two iterate until the candidate matches the reference style.

As no benchmark is dedicated to figure style transfer, we build \texttt{PlotTwin-Bench}. Its references come from two sources: figures replotted by hand from top venues and journals, and existing plotting code that a pipeline enriches into more complex and visually appealing figures. We extract scoring criteria from each reference and score candidates through two channels. The code channel runs an exact check on the attributes where the reference departs from a default plot. The vision channel catches problems that only appear once drawn, such as overlapping text or clipped labels.

Our contributions are:
\begin{itemize}
\item We formulate reference-conditioned style transfer for scientific figures and identify precise style measurement as its core challenge.
\item We propose \emph{Grounded Measurement}, which repurposes coordinate grounding to read each style attribute from the reference with code, and realize it in \texttt{FigMirror}, an agentic Drawer--Reviewer framework.
\item We build \texttt{PlotTwin-Bench}, the first benchmark designed for figure style transfer, with expert-curated figure--code pairs and per-reference scoring criteria that evaluate style from both code and rendered image. \texttt{FigMirror} consistently outperforms prior methods. 
\end{itemize}

%% file: sections/02_related_work.tex
\section{Related Work}

Chart-to-code generation produces plotting code from a chart image. The task grew out of chart understanding, which treated figures as question-answering and derendering targets~\citep{kafle2018dvqa,masry2022chartqa,kantharaj2022chart,liu2023matcha}, and turned generative once multimodal models could emit executable programs. ChartLlama~\citep{han2023chartllama} LoRA-tunes~\citep{hu2021lora} a multimodal LLaMA~\citep{touvron2023llama} to redraw a chart from its image, and successors scale the recipe with larger corpora, code-oriented backbones, and reinforcement learning~\citep{zhao2025chartcoder,tan2025chartmaster}. As general-purpose models matured, training gave way to inference-time self-refinement~\citep{madaan2023self,shinn2023reflexion}: a reviewer compares the render against the reference, and a drawer revises the code until the two agree~\citep{li2025metal,xu2025improved}. Benchmarks co-evolved with the methods, pairing a reference figure with its code and scoring how faithfully a candidate reproduces it~\citep{yang2025chartmimic,wu2025plot2code}.

Computer-use agents complete tasks by operating software through its interface, on the web~\citep{deng2023mind2web,zhou2024webarena} and across full desktops~\citep{xie2024osworld}. Acting on a screen requires knowing where its elements are, so the line invests heavily in GUI grounding: SeeClick~\citep{cheng2024seeclick} pre-trains on grounding data, UGround~\citep{gou2025navigating} scales a universal grounding model on synthetic web pages, OS-ATLAS~\citep{wu2025atlas} builds a grounding corpus across platforms, and agent models such as CogAgent~\citep{hong2024cogagent} and UI-TARS~\citep{qin2025ui} fold grounding into end-to-end action, the predicted coordinates feeding clicks, drags, and keystrokes. The direction is now mainstream: the latest frontier models ship computer use as a built-in capability and benchmark it on OSWorld alongside coding and reasoning~\citep{anthropic2026opus5,openai2026gpt56}.

Neither line reaches our setting. Chart-to-code methods and their benchmarks target reproduction; plotting new data appears at most as an auxiliary test case~\citep{yang2025chartmimic}. No method reads a reference's style attribute by attribute, and no benchmark scores how faithfully a style carries to new data. Computer-use agents train the grounding that such reading needs, but apply it only to on-screen actions. Reference-conditioned style transfer has yet to be studied as a task in its own right.

%% file: sections/03_method.tex
\section{Method}
\label{sec:method}

\begin{figure}[t]
	\centering
	\includegraphics[width=1\linewidth,trim={0.25cm 0.15cm 0cm 0cm},clip]{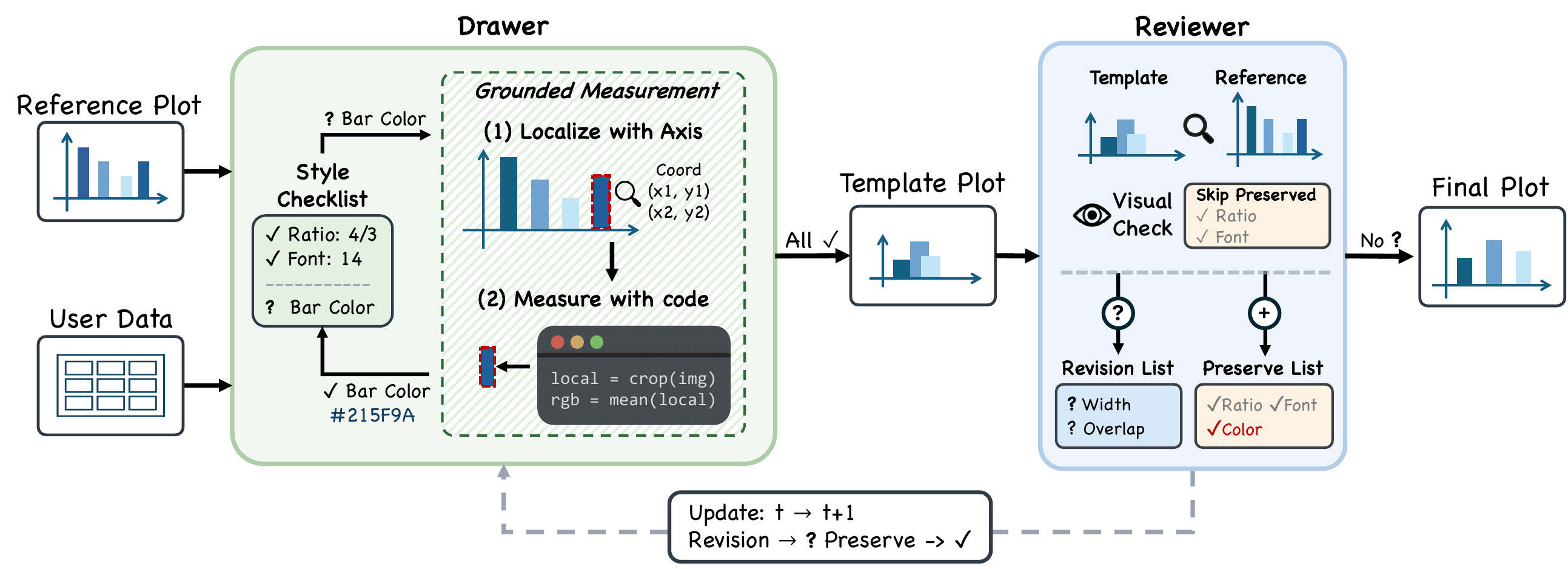}
	\caption{\textbf{The \texttt{FigMirror} pipeline.} \checkmark{} = resolved,
? = open. The Drawer resolves each open style attribute with Grounded
Measurement (locate the element, read its exact value with code) and
renders the user data into a template; the Reviewer checks it visually
against the reference, routing attributes to the Revision or Preserve
List, until no revision remains and the template is final.}
	\label{fig:main-alg}
\end{figure}

\texttt{FigMirror} is an agentic framework for this setting. A Style Checklist organizes which style attributes to inspect on a reference; Grounded Measurement then locates the visual element that carries each attribute and reads its exact value from the reference with code (Section~\ref{sec:grounded-measurement}). A Drawer--Reviewer loop organizes the generation (Section~\ref{sec:drawer-reviewer-loop}): the Drawer resolves every open attribute this way and renders the user data into a template, the Reviewer checks the template against the reference visually and routes attributes into a Revision and a Preserve List, and the lists update the checklist for the next round. The loop ends when the Revision List is empty; Figure~\ref{fig:main-alg} summarizes the process.

\subsection{Grounded Measurement}
\label{sec:grounded-measurement}

A scientific figure shares its anatomy with a graphical interface
(Figure~\ref{fig:gui-plot-comparison}): both are rendered from code into
discrete elements with sharp boundaries and flat fills. On such pixels,
general-purpose models are trained to ground a referred element~\citep{cheng2024seeclick,gou2025navigating,xie2024osworld}: given an
image $I$ and an element $e$, such as a button or a menu entry, they
return its image coordinates $(x, y) = g(I, e)$. An axis tick or a legend
entry grounds just as well, so we reuse this coordinate interface on the
reference figure.

Grounded Measurement builds on the reused interface.
\textit{\textbf{(1) Localize.}} For a style attribute $a$, the model is queried
for the coordinates that delimit the visual element carrying $a$, the
two corners $(x_1, y_1)$ and $(x_2, y_2)$ that span a region $r_a$ of
the reference. \textit{\textbf{(2) Measure.}} A short program $\rho_a$ computes
the attribute value $v_a$ from that region,
\begin{equation}
  v_a = \rho_a\big(I[r_a]\big),
\end{equation}
where $I[r_a]$ denotes the local crop used to measure $a$.
The two steps appear as step (1) and (2) in the Drawer part of Figure~\ref{fig:main-alg}.

We then observe that the measurement programs are supplied just as
readily. Code is the modality frontier models optimize hardest~\citep{wang2024executable}, and the
few lines that read one attribute sit well within this capability. The
lines differ by attribute, and the model fits them to what it sees: a
mean over a flat fill recovers a bar's color, a thin anti-aliased
stroke asks instead for the modal pixel, a run test along a stroke
tells dashed from solid. One primitive thus spans the attribute space,
from continuous values such as colors, widths, and aspect ratios to
discrete ones such as the presence of grid lines.
\begin{wrapfigure}[17]{r}{0.47\linewidth}
  \centering
  \includegraphics[width=\linewidth,trim={4pt 14pt 14pt 8pt},clip]{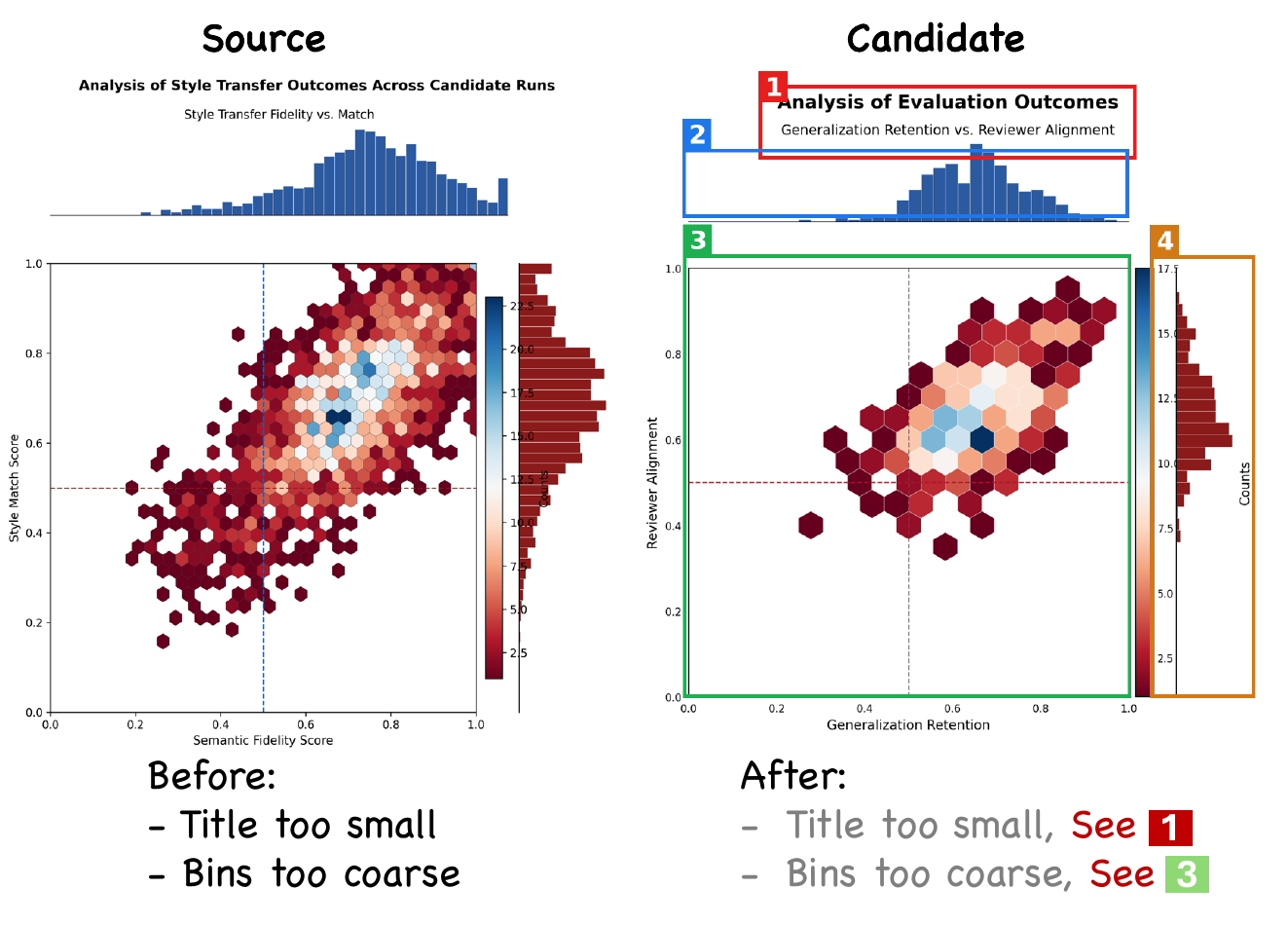}
  \vspace{-0.65cm}
  \caption{\textbf{Grounded Reviewer feedback.}
  Plain text leaves the repair target ambiguous; a note tied to a marked
  region names it exactly, so the next Drawer revises the local mismatch
  and leaves matched attributes untouched.}
  \label{fig:review-feedback}
  \vspace{-0.55in}
\end{wrapfigure}
The same coordinate interface also supports review. After a draft is
rendered, the model can mark the region where a style attribute was
measured or applied incorrectly. This turns a vague visual critique into
a local repair target.

\subsection{The Drawer--Reviewer Loop}
\label{sec:drawer-reviewer-loop}

A single measurement reads one attribute; a figure has dozens. The
Drawer therefore maintains a Style Checklist, composed of an open set
$\mathcal{O}$ and a resolved set $\mathcal{R}$. The checklist is
initialized under prompt guidance, where the Drawer scans the reference
against a predefined attribute list and collects the attributes the
figure exhibits, all open at the start. Each open attribute is then
resolved by Grounded Measurement and moved to $\mathcal{R}$ with its
value $v_a$. Once $\mathcal{O}$ is empty, the Drawer writes plotting
code for the user data using the resolved style values and renders a
draft figure.

The Reviewer gives this draft a fresh visual check. It receives a
side-by-side view of the reference and the draft for global comparison,
together with separate high-resolution views for local details. The
side-by-side view exposes layout and balance errors. The high-resolution
views expose small failures such as text overlap or misplaced guides. As
illustrated in Figure~\ref{fig:review-feedback}, the Reviewer returns
feedback items $(b_j, n_j)$, where $b_j$ is a marked region and $n_j$ is
a short note. For example, a sentence such as ``the label is too close''
leaves the Drawer guessing which label and which mark are involved. A
marked region tells the Drawer exactly what the critique refers to.

The loop alternates a stateful Drawer and a stateless Reviewer. In each
iteration, the Drawer edits the previous code using two lists from the
last review. The Preserve List contains accepted style choices and grows
monotonically across iterations. The Revision List contains localized mismatches and returns the corresponding attributes to $\mathcal{O}$.
The Reviewer re-audits the full draft in every round. This separation lets the
Drawer preserve useful construction state while the Reviewer judges only
the rendered result. The loop stops when a review produces an empty
Revision List, and the current draft becomes the final figure.

\textbf{Implementation.} \texttt{FigMirror} is packaged as a \textit{skill}, a
self-contained instruction bundle that an agentic coding harness loads
at run time. The skill carries the full procedure (the Style Checklist,
Grounded Measurement, and the Drawer--Reviewer loop); the harness
supplies the model, code execution, and image access. This separation
keeps \texttt{FigMirror} portable across harnesses.

%% file: sections/04_experiments.tex
\section{Benchmark and Evaluation}
\label{sec:benchmark-evaluation}

\subsection{Benchmark Construction}

\textbf{Data sources.}
Existing chart-to-code benchmarks mostly focus on reproduction, 
so we build \texttt{PlotTwin-Bench}. It targets real-world use: a reference
figure is worth mirroring when its visual design is worth reusing and its structure is
easier to specify with an image than with a text prompt. We therefore
select references that are visually appealing and structurally rich. We
draw on two sources. The first is hand-curated. We
select complex, visually appealing figures from papers in top venues, then
replot each figure by hand to obtain an aligned figure--code pair. This
source contains 50 figures. The second source
scales this up. We start from existing plotting code from
ChartMimic~\citep{yang2025chartmimic} and use an LLM-based rewriting
pipeline to produce more complex and visually appealing references
(details in Appendix~\ref{app:reference-enrichment}). A filter then
removes the figures that fall
short in either property, leaving 350. Together the two sources give
400 references across 12 chart types, including grouped bars,
multi-panel line plots, and heatmaps. Examples of this enrichment
appear in Appendix~\ref{app:reference-enrichment}.
Figure~\ref{fig:type-distribution} shows the five most common chart
types in each source. Appendix~\ref{app:benchmark-composition} reports
the full composition.

\textbf{Style-transfer task.}
Each task pairs a reference figure with target data and asks the model
to visualize the data in the reference's style. To construct the target
data, a generator sees only the reference image, invents a plausible
scientific story in a different domain, and derives a dataset from that
story. It then changes one to three aspects of the dataset while keeping
it compatible with the reference's chart and panel structure. The
generator performs a final check for structural consistency and
numerical plausibility; Appendix~\ref{app:augment} gives the full
procedure.

\begin{figure}[!htbp]
  \centering
  \begin{subfigure}[b]{0.47\linewidth}
    \centering
    \includegraphics[width=\linewidth]{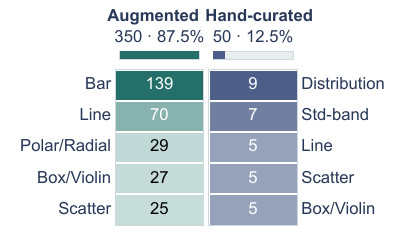}
    \caption{}
    \label{fig:type-distribution}
  \end{subfigure}
  \hspace{-0.010\linewidth}
  \begin{subfigure}[b]{0.38\linewidth}
    \centering
    \includegraphics[width=\linewidth]{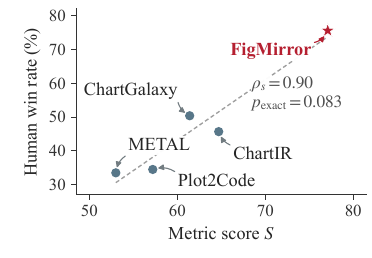}
    \caption{}
    \label{fig:human-metric-alignment}
  \end{subfigure}
  \vspace{-0.45em}
  \caption{\textbf{Benchmark composition and evaluator alignment.}
  (a)~Five most common chart types in each source.
  (b)~Automated score $S$ versus human win rate across the five methods.}
  \label{fig:benchmark-summary}
\end{figure}

\subsection{Evaluation}

Direct VLM scoring tends to be lenient for scientific figure style
transfer~\citep{zheng2023judging}. Most figures share standard axes, marks, and labels, so a
holistic comparison can underweight the few choices that distinguish a
reference and give visibly different candidates similar scores. Our
evaluator focuses on these reference-specific choices through two
complementary channels. The code channel identifies departures from an
average plotting style, while the vision channel finds important
mismatches in the rendered figure.

The code channel scores the attributes that set a reference apart from
an average figure. We use the Matplotlib~\citep{hunter2007matplotlib} defaults as a fixed and
reproducible stand-in for this average. For a style attribute $a$, let $v_a$ be its reference value, $d_a$ its
default value, and $\hat v_a$ its candidate value. PlotTwin-Bench pairs
every reference with its code. The evaluator therefore reads $v_a$ from
the reference code and $\hat v_a$ from the candidate code, and both
values are exact. During generation, the method sees only the rendered
reference and measures from pixels
(Section~\ref{sec:grounded-measurement}). The scored set is
\begin{equation}
  \Delta(I) = \{a \mid v_a \neq d_a\}.
\end{equation}
For each $a\in\Delta(I)$, the candidate receives full credit when
$\hat v_a$ matches $v_a$, zero credit when it is no closer to $v_a$ than
$d_a$, and proportional credit in between. We use exact agreement for
categorical choices and normalized distance for continuous values. If
$q_a\in[0,1]$ denotes this credit, the code score is
\begin{equation}
  S_{\mathrm{code}} =
  \frac{100}{|\Delta(I)|}\sum_{a\in\Delta(I)} q_a.
\end{equation}

Code does not capture every distinctive choice. Layout, spacing,
alignment, and visual hierarchy are often read more clearly from the
rendered figure. The vision channel uses a VLM to identify important
reference choices that the candidate failed to reproduce and records
them as evidence-grounded visual defects. Each finding must cite
visible evidence through bounding boxes in the reference and candidate
images and, when applicable, the relevant code lines. The vision score
$S_{\mathrm{vision}}$ starts at $100$ and deducts $5$, $10$, or $25$
points for each minor, major, or critical defect. The evidence
requirement constrains unsupported findings and makes every deduction
auditable.

The two scores cover complementary evidence on the same scale. We
combine them as
\begin{equation}
  S = 0.35 S_{\mathrm{code}} + 0.65 S_{\mathrm{vision}}.
\end{equation}
The weights are fixed a priori: readers judge the rendered figure, so
the vision channel receives the larger weight. We use the same weights
in all experiments and report both channel scores alongside $S$. A human study suggests $S$ aligns with human
judgment (Figure~\ref{fig:human-metric-alignment});
Appendix~\ref{app:human-study} reports the protocol and statistics.

\section{Experiments}

\subsection{Experimental Setup}

We evaluate on 150 of the 400 references: all 50 hand-curated figures
and 100 sampled at random from the augmented source; the exact subset
is included in the benchmark release. Every method returns a
self-contained plotting script, scored with the evaluator of
Section~\ref{sec:benchmark-evaluation}. Both the evaluator and every
method's underlying model are GPT-5.5 at x-high reasoning effort.
\texttt{FigMirror} runs as a skill in the Codex harness
(Section~\ref{sec:method}). Appendix~\ref{app:experiment-setup} reports
the rendering environment, harness configuration, and iteration budget;
Appendix~\ref{app:prompts} gives the prompt bundle.

\textbf{Baselines.} We compare \texttt{FigMirror} with four external
chart-to-code baselines: Plot2Code~\citep{wu2025plot2code}, a one-shot
chart-to-code method; METAL~\citep{li2025metal}, an iterative feedback
method for chart code generation;
ChartGalaxy-Prompt~\citep{li2025chartgalaxy} (hereafter
ChartGalaxy), the prompt-based code-generation recipe released with
the dataset of the same name; and
ChartIR~\citep{xu2025improved}, a method that repairs generated code
over successive rounds. We extract each reference's scoring criteria
once and share them across all methods.

\subsection{Main Results}

Table~\ref{tab:main-results} reports the main style-transfer
comparison on \texttt{PlotTwin-Bench}, with the hand-curated and augmented
splits shown separately. \texttt{FigMirror} has the best combined score on
both, leading the strongest baseline, ChartIR, by 11.4 points on the
hand-curated split (72.7 versus 61.3) and by 6.1 on the augmented
split (76.4 versus 70.3). It also leads in both channels on both
splits, so the lead does not depend on the weights in $S$. Iteration
alone does not order the
baselines. ChartIR improves over the one-shot Plot2Code, while METAL,
also iterative, stays at Plot2Code's level. Its critic is built for
reproduction, and once the data differ, whole-image comparison pulls
the draft toward the reference's data rather than its style.

The two margins trace back to where each split's styles come from.
Augmented references are rewritten from existing plotting code, and a
strong code model recovers much of their style unaided. The three
leading methods all exceed 80 in the code channel, leaving the vision
channel to separate them (9.2 points). Hand-curated references carry
choices made by paper authors, which must be read from the image. On
this split \texttt{FigMirror} leads in both channels, by 5.8 points in code
and 13.3 in vision. These are the references the benchmark is built
around, designs worth reusing that the model cannot guess.

\begin{figure}[!t]
  \centering
  \includegraphics[width=\linewidth]{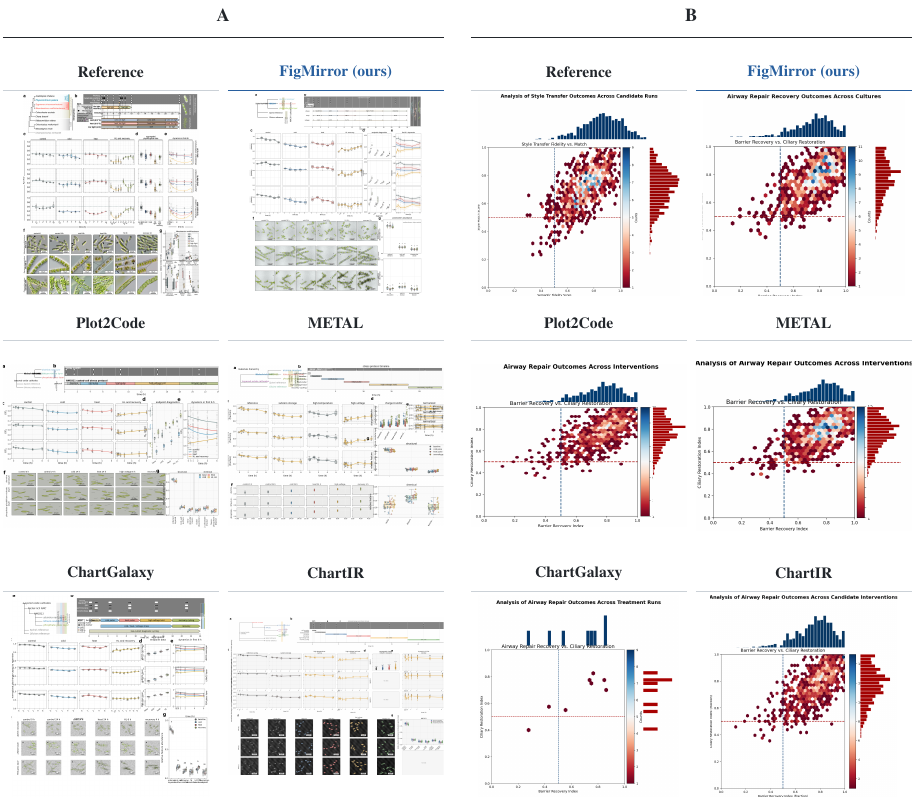}
  \vspace{-0.35em}
  \caption{\textbf{Style transfer on two references.}
  Each group places \texttt{FigMirror} beside the reference, followed by four
  baselines. In A, \texttt{FigMirror} reproduces the panel hierarchy and relative
  spacing of a dense composition. In B, it matches the joint hexbin layout,
  marginal histograms, threshold lines, color scale, and typography.}
  \label{fig:qualitative}
\end{figure}

\input{tables/main_results}

Figure~\ref{fig:qualitative} compares five methods on two references.
Within each group, every method receives the same reference and target
data.

\subsection{Ablations}

Table~\ref{tab:ablation} removes the two parts of \texttt{FigMirror} we expect
to matter most, then the skill as a whole. The first ablation removes
the Reviewer and keeps the Drawer's drafts and self-checks. The second
removes Grounded Measurement from the feedback loop. The third, naked
Codex, drops the skill and prompts the same harness directly. All
three use the style-transfer setting.

Each ablation lowers the combined score, and the loss grows with what
is removed. Removing the Reviewer costs 2.1 points. Mismatches that
get past the Drawer's self-checks reach the final draft uncorrected,
and both channels drop about two points. Removing Grounded
Measurement costs 3.7. The Drawer estimates attribute values by eye,
the Reviewer still boxes the resulting mismatches, but the corrections
are also made by eye. The surviving drift costs 7.2 points in the
vision channel, while the code channel edges up. Naked Codex costs
11.7. It drafts by eye alone and never measures the
reference, so both channels fall and the score drops to the level of
the external baselines.

\begin{figure}[!t]
  \centering
  \includegraphics[width=\linewidth]{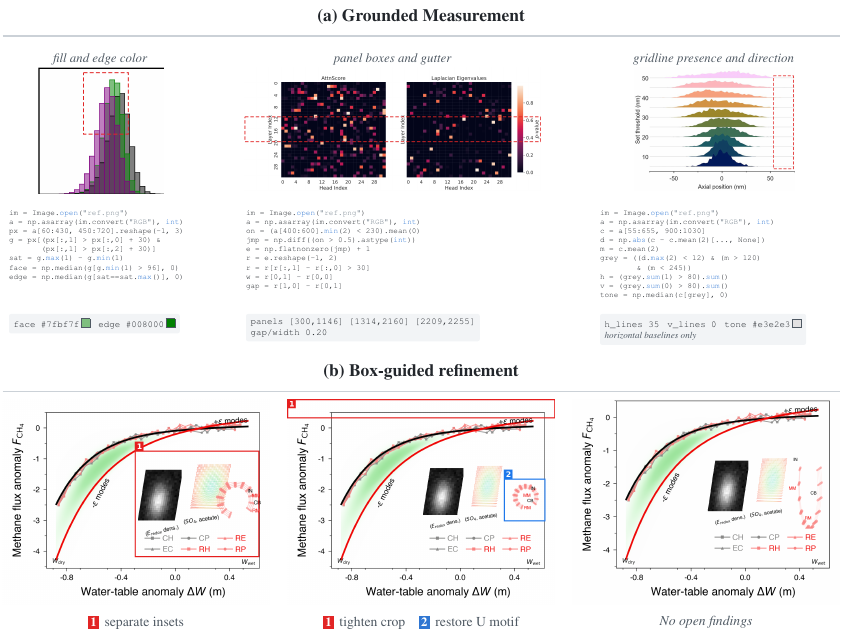}
  \vspace{-0.35em}
  \caption{\textbf{Grounded Measurement and box-guided refinement.}
  (a)~Three measurements from separate runs: each dashed box marks the
  probed region on a reference; below it, the probe code and its
  returned values. (b)~One transfer run over three iterations; labels
  name the Drawer edit applied in the next column, and unboxed
  attributes stay in the Preserve List.}
  \label{fig:mechanism-case-study}
\end{figure}

\input{tables/ablation}

\subsection{Iteration Budget}

We test whether additional Drawer and Reviewer rounds improve style
transfer under a fixed Codex configuration
(Table~\ref{tab:max-iteration}).

\input{tables/max_iteration}

The combined score rises with the budget, from 71.8 at one iteration
to 72.7 at three and 74.3 at five. At three iterations, the vision
score increases by 1.4 points while the code score remains essentially unchanged.
Five iterations improve both channels. We use three in the main
experiments to limit inference cost.

\subsection{Mechanism-Level Case Study}

Figure~\ref{fig:mechanism-case-study}(b) follows one transfer run
through three iterations. Each review binds a mismatch to a box, and
the next iteration edits inside the boxes while unboxed attributes,
held in the Preserve List, carry over unchanged.

The run makes the localization argument of
Section~\ref{sec:drawer-reviewer-loop} concrete. The first review marks
the overlapping insets, and iteration~1 separates them: the box turns a
composition-wide search into a local edit.

Because the Reviewer is stateless, each round re-audits the full draft,
and subtler mismatches surface once dominant ones are cleared. With the
insets separated, the next review boxes the broad top margin and the
compressed U-shaped motif; iteration~2 fixes both, and the final review
returns no boxes. Every boxed defect is resolved in the following
iteration, and no repair disturbs an attribute already accepted.
Panel (a) shows the values that feed these repairs: each probe returns
the exact value the Drawer commits to the checklist.

%% file: tables/main_results.tex
\begin{table*}[t]
  \centering
  \caption{\textbf{Main results on scientific figure style transfer.}
  We report the code score $S_{\mathrm{code}}$, the vision score
  $S_{\mathrm{vision}}$, and the combined score $S$ on the two reference
  sources of \texttt{PlotTwin-Bench}: hand-curated figures and augmented figures.
  All methods use GPT-5.5.}
  \label{tab:main-results}
  \small
  \setlength{\tabcolsep}{4.5pt}
  \begin{tabular}{@{}lcccccc@{}}
    \toprule
    & \multicolumn{3}{c}{Hand-curated} & \multicolumn{3}{c}{Augmented} \\
    \cmidrule(lr){2-4}\cmidrule(lr){5-7}
    Method
      & $S_{\mathrm{code}}$ & $S_{\mathrm{vision}}$ & $S$
      & $S_{\mathrm{code}}$ & $S_{\mathrm{vision}}$ & $S$ \\
    \midrule
    {\bf\texttt{FigMirror}}
      & \textbf{64.4} & \textbf{77.2} & \textbf{72.7}
      & \textbf{84.0} & \textbf{72.3} & \textbf{76.4} \\
    ChartIR~\citep{xu2025improved}
      & 56.6 & 63.9 & 61.3
      & 83.6 & 63.1 & 70.3 \\
    ChartGalaxy~\citep{li2025chartgalaxy}
      & 58.6 & 58.4 & 58.5
      & 80.7 & 58.1 & 66.0 \\
    Plot2Code~\citep{wu2025plot2code}
      & 53.4 & 56.8 & 55.6
      & 64.9 & 43.0 & 50.7 \\
    METAL~\citep{li2025metal}
      & 53.7 & 50.5 & 51.6
      & 66.2 & 43.5 & 51.5 \\
    \bottomrule
  \end{tabular}
\end{table*}

%% file: tables/ablation.tex
\begin{table}[t]
  \centering
  \caption{\textbf{Ablations on style transfer.} We implement each
  variant in the Codex harness by editing the corresponding skill
  prompts and removing the relevant tool access. The no-Reviewer
  variant retains the Drawer's self-checks.}
  \label{tab:ablation}
  \small
  \begin{tabular}{lccc}
    \toprule
    Variant & $S_{\mathrm{code}}$ & $S_{\mathrm{vision}}$ & $S$ \\
    \midrule
    \texttt{FigMirror} & 64.4 & 77.2 & 72.7 \\
    w/o Reviewer & 62.7 & 74.9 & 70.6 \\
    w/o Grounded Measurement & 67.4 & 70.0 & 69.0 \\
    w/o \texttt{FigMirror} skill (naked Codex) & 60.4 & 61.4 & 61.0 \\
    \bottomrule
  \end{tabular}
  \vspace{-1.5em}
\end{table}

%% file: tables/max_iteration.tex
\begin{wraptable}[7]{r}{0.42\linewidth}
  \vspace{-0.8\baselineskip}
  \centering
  \caption{\textbf{Iteration budget in the Codex setting.} All rows
  share the same model, data, and evaluation.}
  \label{tab:max-iteration}
  \small
  \begin{tabular}{lccc}
    \toprule
    Max iterations & $S_{\mathrm{code}}$ & $S_{\mathrm{vision}}$ & $S$ \\
    \midrule
    1 & 64.5 & 75.8 & 71.8 \\
    3 \textit{(default)} & 64.4 & 77.2 & 72.7 \\
    5 & 67.0 & 78.3 & 74.3 \\
    \bottomrule
  \end{tabular}
\end{wraptable}

%% file: sections/07_ethics.tex
A scientific figure carries an argument through data and the way that
data is presented. \texttt{FigMirror} helps a user transfer a presentation style
after the user has chosen a suitable reference; it does not decide
whether that presentation fits the data. The user stays responsible for
the substance of the figure, including the axis scales, labels, units,
legends, and captions, and for checking that the visual encoding
supports the underlying claim.

Our benchmark is built from licensed sources. In constructing it, we
select reference figures from sources with licenses that permit research
use and retain their provenance during curation. When applying \texttt{FigMirror}
outside the benchmark, users should respect the license and attribution
terms of any reference figure and treat the generated figure as an
auditable draft rather than publication-ready evidence.

%% file: appendix/appendix.tex
\section{Limitations}
\label{app:limitations}
\input{sections/06_limitations}

\section{Additional Implementation Details}

\subsection{Benchmark Composition}
\label{app:benchmark-composition}

\begin{figure}[H]
  \centering
  \includegraphics[width=\linewidth]{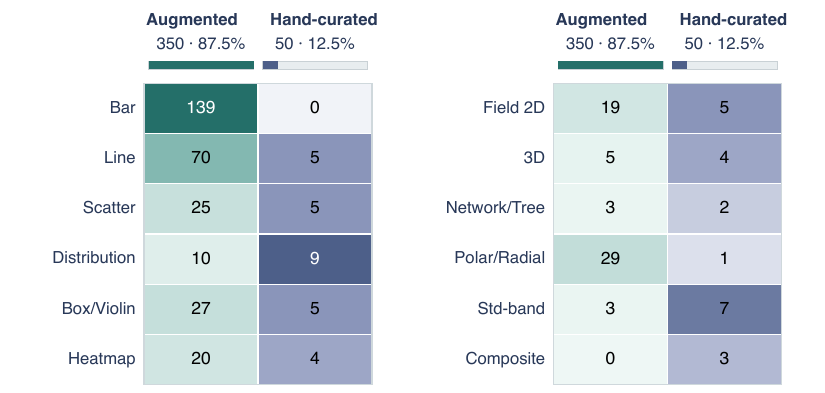}
  \vspace{-0.35em}
  \caption{\textbf{Full chart-type distribution.}
  Counts for all 12 chart types in the augmented and hand-curated
  sources.}
  \label{fig:type-distribution-full}
\end{figure}

Figure~\ref{fig:type-distribution-full} reports the complete chart-type
distribution for the augmented and hand-curated sources.

\subsection{Reference Enrichment Pipeline}
\label{app:reference-enrichment}

\begin{figure*}[p]
  \centering
  \makebox[0.44\textwidth][c]{\textbf{Seed}}%
  \hspace{0.015\textwidth}%
  \makebox[0.44\textwidth][c]{\textbf{Augmented}}
  \par\vspace{0.2em}
  \makebox[0.44\textwidth][c]{%
    \includegraphics[width=0.44\textwidth,height=0.28\textwidth,keepaspectratio]{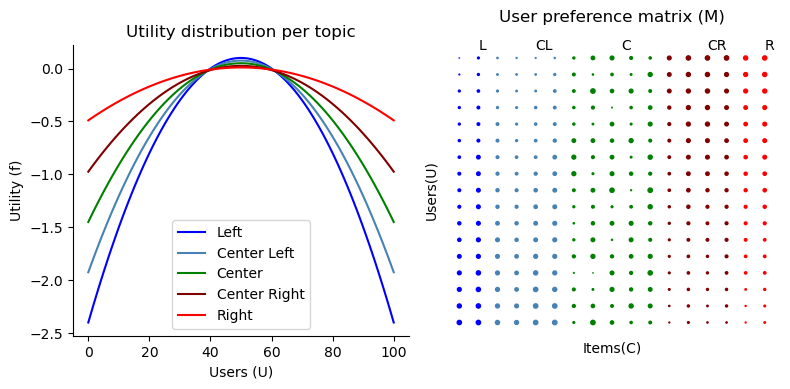}}%
  \hspace{0.015\textwidth}%
  \makebox[0.44\textwidth][c]{%
    \includegraphics[width=0.44\textwidth,height=0.28\textwidth,keepaspectratio]{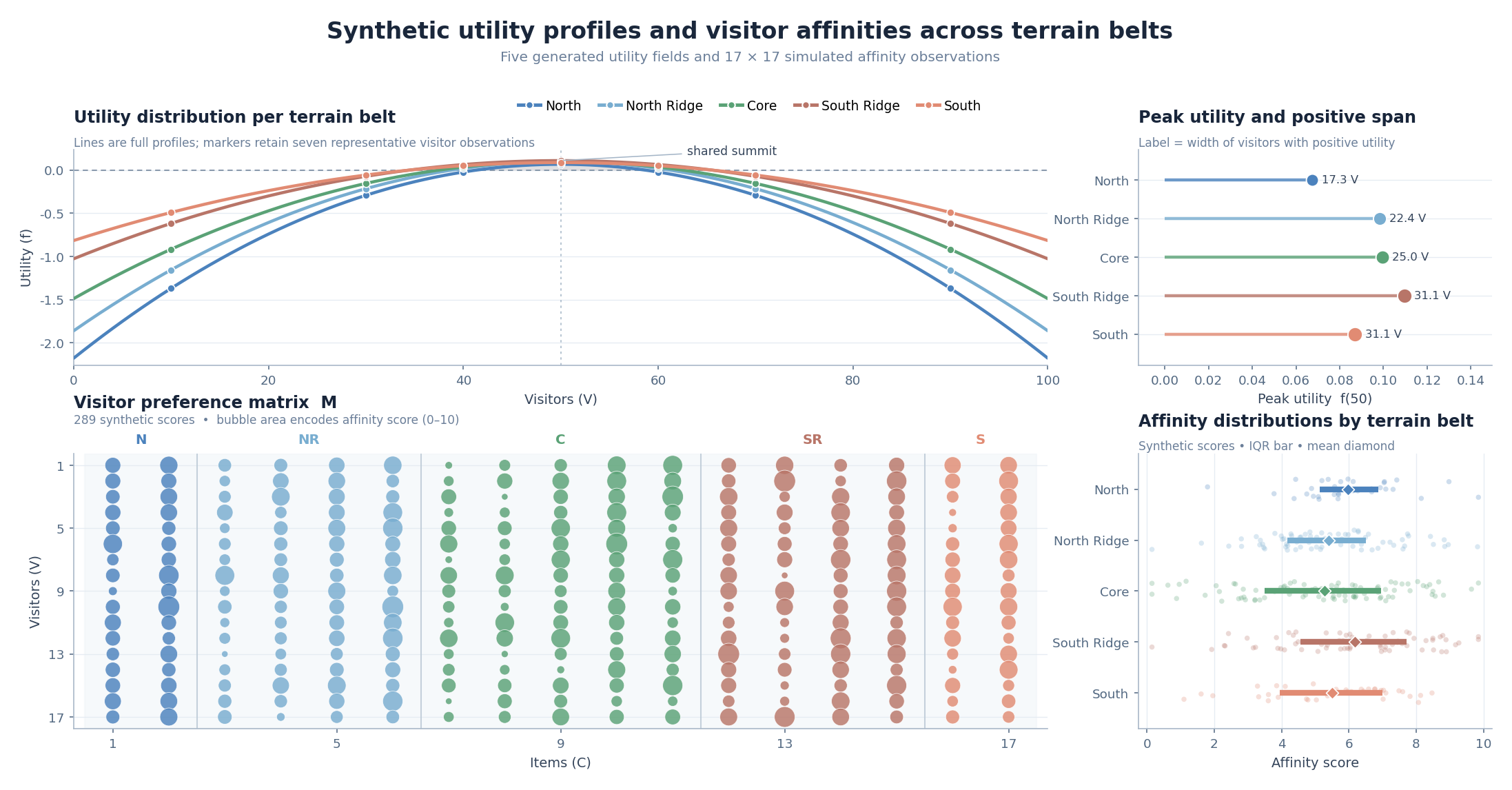}}
  \par\vspace{-0.45em}
  {\small (a)}
  \par\vspace{0.1em}
  \makebox[0.44\textwidth][c]{%
    \includegraphics[width=0.44\textwidth,height=0.28\textwidth,keepaspectratio]{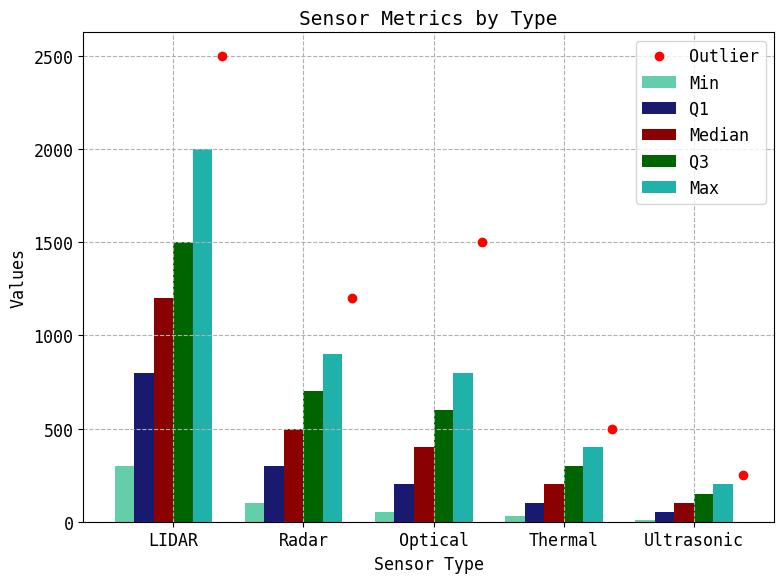}}%
  \hspace{0.015\textwidth}%
  \makebox[0.44\textwidth][c]{%
    \includegraphics[width=0.44\textwidth,height=0.28\textwidth,keepaspectratio]{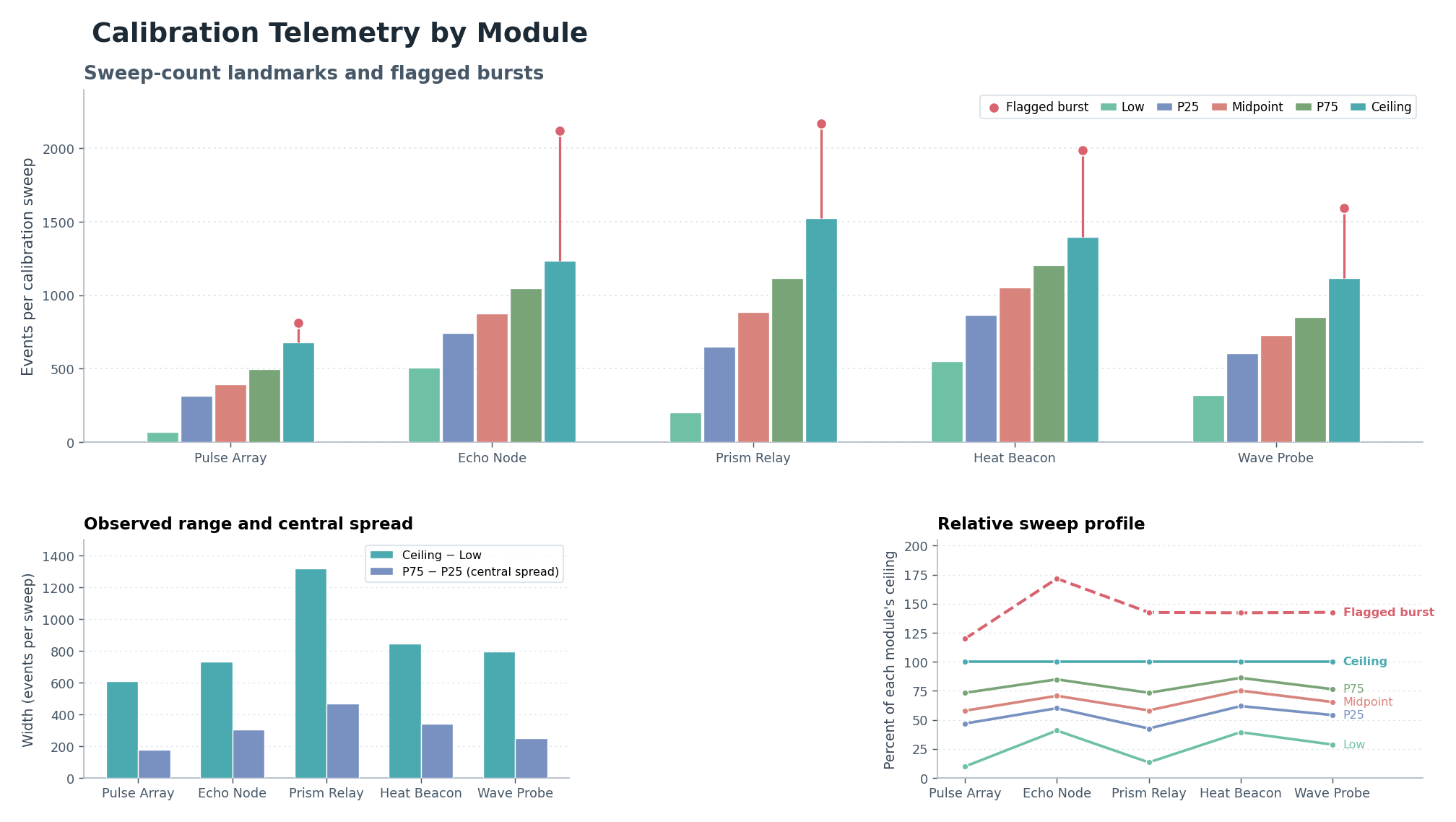}}
  \par\vspace{-0.45em}
  {\small (b)}
  \par\vspace{0.1em}
  \makebox[0.44\textwidth][c]{%
    \includegraphics[width=0.44\textwidth,height=0.28\textwidth,keepaspectratio]{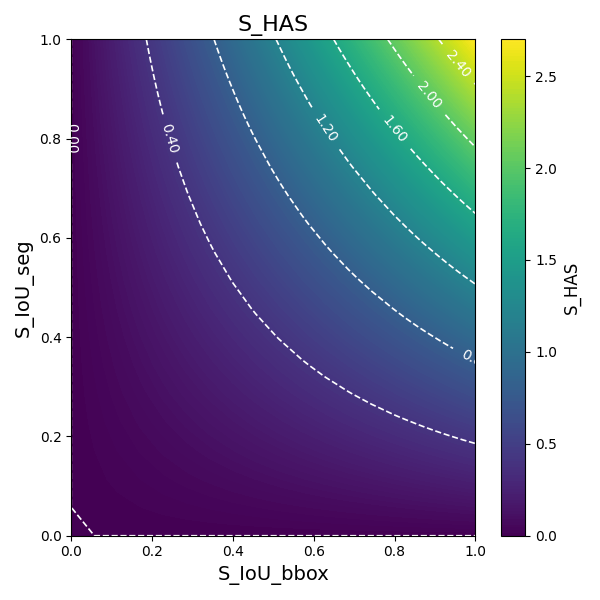}}%
  \hspace{0.015\textwidth}%
  \makebox[0.44\textwidth][c]{%
    \includegraphics[width=0.44\textwidth,height=0.28\textwidth,keepaspectratio]{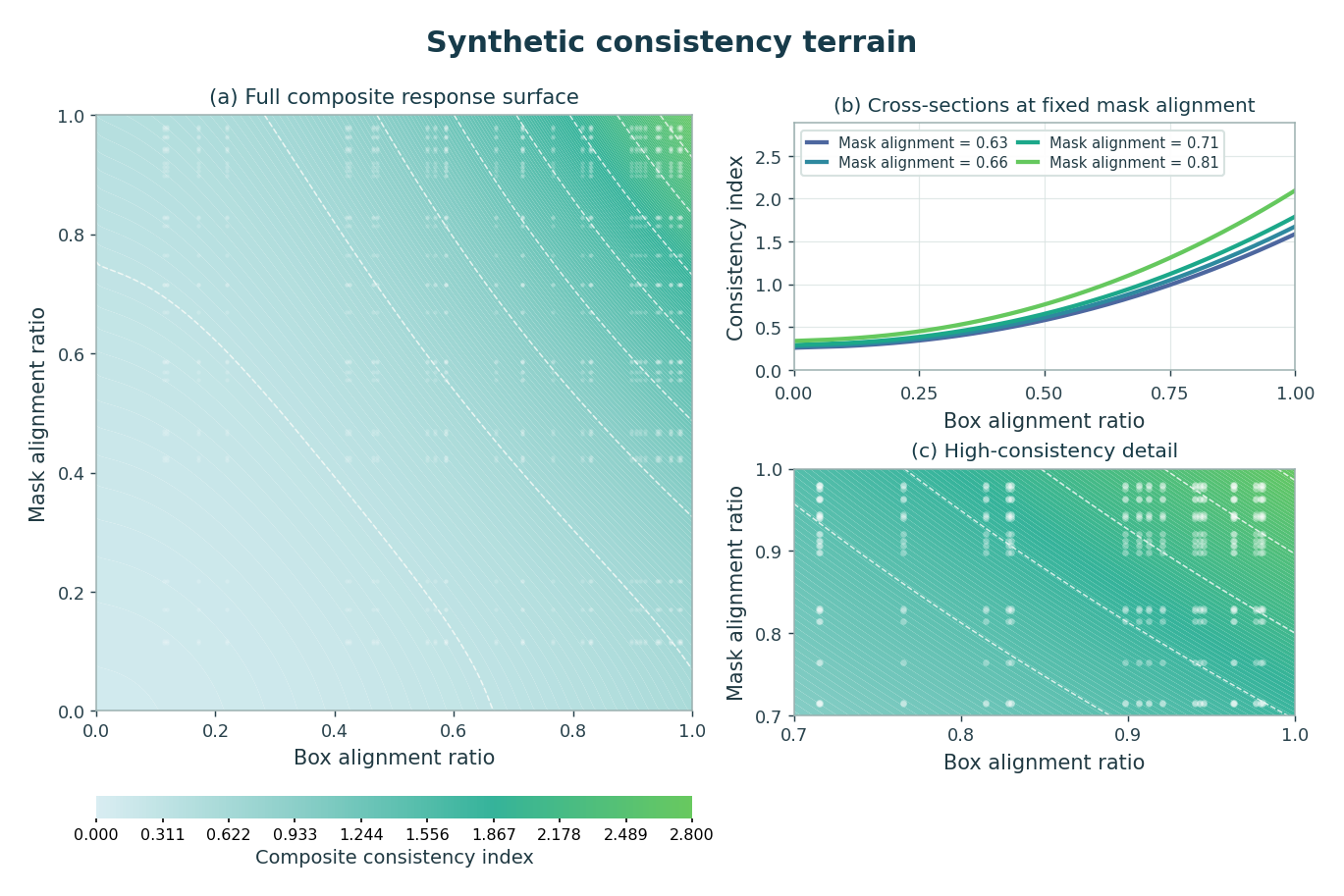}}
  \par\vspace{-0.45em}
  {\small (c)}
  \par\vspace{0.1em}
  \makebox[0.44\textwidth][c]{%
    \includegraphics[width=0.44\textwidth,height=0.28\textwidth,keepaspectratio]{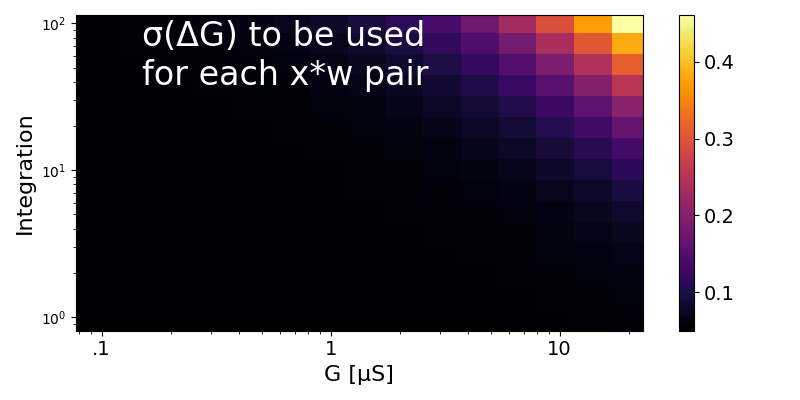}}%
  \hspace{0.015\textwidth}%
  \makebox[0.44\textwidth][c]{%
    \includegraphics[width=0.44\textwidth,height=0.28\textwidth,keepaspectratio]{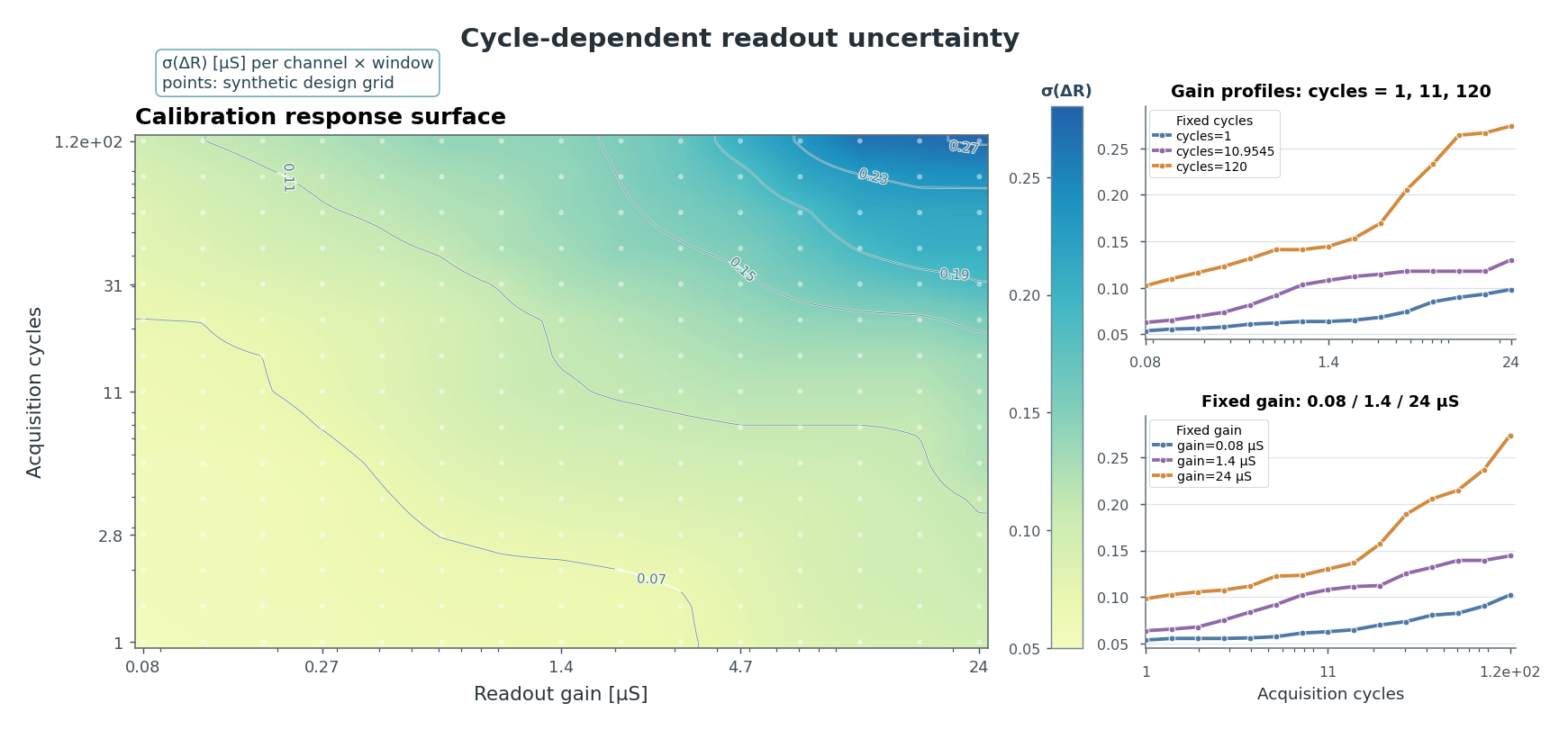}}
  \par\vspace{-0.45em}
  {\small (d)}
  \par\vspace{0.1em}
  \makebox[0.44\textwidth][c]{%
    \includegraphics[width=0.44\textwidth,height=0.28\textwidth,keepaspectratio]{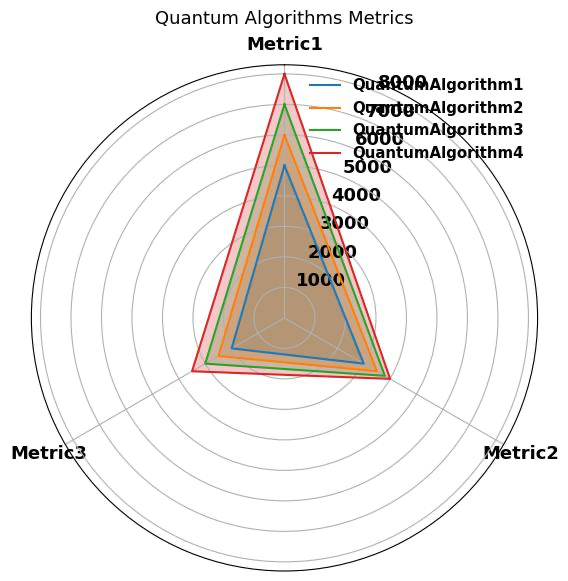}}%
  \hspace{0.015\textwidth}%
  \makebox[0.44\textwidth][c]{%
    \includegraphics[width=0.44\textwidth,height=0.28\textwidth,keepaspectratio]{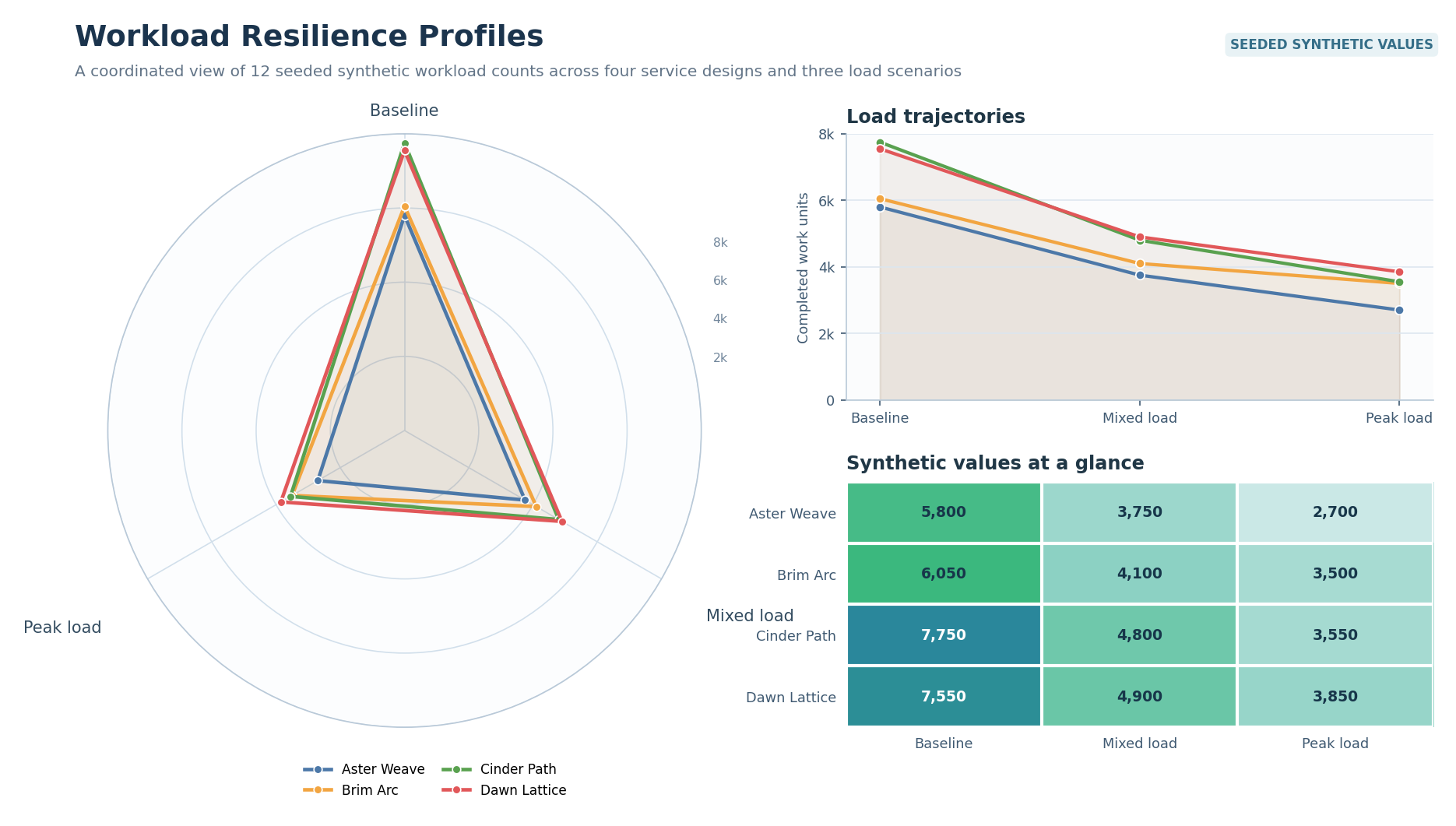}}
  \par\vspace{-0.45em}
  {\small (e)}
  \vspace{-0.35em}
  \caption{\textbf{Reference enrichment examples.}
  Each row places the seed figure on the left and its augmented reference on
  the right. The augmented versions add coordinated panels and retain the
  seed's principal marks and encodings.}
  \label{fig:reference-enrichment-examples}
\end{figure*}

We enlarge the reference pool from ChartMimic plotting code using an
LLM-based rewriting pipeline. The pipeline rewrites each plotting script
toward higher structural complexity and visual polish, and retains
references that satisfy both criteria. Figure~\ref{fig:reference-enrichment-examples}
shows five retained rewrites.

\subsection{Transfer-Data Construction}
\label{app:augment}

\paragraph{Story-first generation.}
For each reference, the generator receives only the reference image. It
reads the figure's chart types, panel structure, major regions, and
panel hierarchy, then invents a plausible scientific story in a
different domain. From this story it defines the variables, conditions,
measurements, repetitions, and uncertainty, and writes one CSV whose
fields cover the visual roles in the reference.

\paragraph{Variation and visual compatibility.}
The generator chooses one to three dimensions from
Table~\ref{tab:augment} to vary the synthesized data. Because changing
the scientific domain already changes labels and axis meanings,
\texttt{label\_domain\_swap} and \texttt{axis\_semantics\_swap} cannot
serve as the main variation by themselves. At least one selected
dimension must instead change the data schema, cardinality, density,
scale, or the range and sign of its values. Throughout, the reference
stays a viable visual template: its main chart types, macro layout,
reading order, and
panel hierarchy are preserved. A regular grid changes its row or column
count only when the new study calls for it and stays a complete grid
such as $2\times2$ or $2\times4$. A heterogeneous figure keeps its macro
regions while at least one local block changes its panel allocation,
grouping, or schema.

\input{tables/augment_ops}

We define each dimension below.

\emph{Data shape (6).}
\texttt{series\_count}: number of distinct series, lines, or groups.
\texttt{point\_density}: rows per series along the independent axis.
\texttt{category\_cardinality}: number of categories on a discrete axis.
\texttt{matrix\_dimension}: row $\times$ column count of a matrix or
heatmap. \texttt{panel\_count}: number of subplots in a multi-panel
figure. \texttt{per\_series\_density\_imbalance}: density ratio across
series.

\emph{Axis and scale (5).}
\texttt{scale\_type}: linear, log, or symlog.
\texttt{x\_axis\_type}: numeric, time, or categorical.
\texttt{axis\_units\_normalization}: a unit or normalization change
(e.g., seconds to milliseconds, raw counts to rates).
\texttt{value\_sign\_polarity}: positive-only versus mixed values that
cross zero. \texttt{dual\_scale\_requirement}: single versus dual y-axis.

\emph{Semantic and categorical (5).}
\texttt{label\_domain\_swap}: replace a category set with a same-size set
from a different domain. \texttt{label\_length\_inflation}: short labels
to long labels. \texttt{ordering\_principle\_swap}: sort by $x$ versus
sort by $y$. \texttt{hierarchy\_introduction}: flat versus grouped
(super- and sub-category) structure. \texttt{axis\_semantics\_swap}:
replace the axis meaning (e.g., cities $\times$ months to genes $\times$
conditions).

\paragraph{Output and checks.}
The generator writes one CSV with descriptive column names and units,
using panel or record identifiers when the file contains several visual
roles. Before finishing, it checks the table for consistent row width,
complete role coverage, finite values, and plausible numerical
relationships. The pipeline then verifies that the output file is
present and well formed. A missing or malformed CSV may be repaired
once; no additional candidates are generated or ranked.

\subsection{Evaluation Protocol Details}
\label{app:evaluation}

Scoring runs after generation. The evaluator reads the source code and
image, the candidate code and image, and the render record. It does not
call any method or rerender a figure, so a method never sees the scoring
criteria while it draws. Each reference receives a code score
$S_{\mathrm{code}}$ and a vision score $S_{\mathrm{vision}}$, which stay
separately auditable.

\paragraph{Code channel.}
The Matplotlib defaults define a fixed stand-in for an average plot: a four-side box,
outward ticks, no grid, the default color cycle, a boxed legend, and the
default aspect, line width, and sans-serif type. For each style
attribute $a$, we read the default value $d_a$, the reference value
$v_a$, and the candidate value $\hat v_a$ from code. We retain the
attribute only when $v_a \neq d_a$, giving the reference-specific set
$\Delta(I)=\{a\mid v_a\neq d_a\}$. The per-attribute credit $q_a$
measures how far the candidate moves from the default toward the
reference: $1$ when it reproduces the reference value, $0$ when it stays
at the default or moves the wrong way, and a graded value in between. Categorical
attributes (grid, removed spines, tick direction, serif, legend frame)
use direct agreement; continuous attributes (palette, background, line
width, aspect) use a relative distance, with colors compared by CIEDE2000.
The code score is
\[
  S_{\mathrm{code}} =
  \frac{100}{|\Delta(I)|}\sum_{a\in\Delta(I)}q_a.
\]

\paragraph{Vision channel.}
This channel judges the distinctive choices that code cannot see, such
as layout, spacing, alignment, and visual hierarchy. A VLM compares the
reference and candidate images and lists the choices the candidate
failed to reproduce. Each listed miss must cite visible evidence:
bounding boxes in the reference and candidate images, and, when the
cause is traceable to the program, the relevant code lines. The
evaluator rates each miss minor, major, or critical, with penalties of
$5$, $10$, and $25$. The score starts at $100$ and subtracts the
penalties, floored at $0$,
\[
  S_{\mathrm{vision}} = \max\!\left(0,\; 100 - \sum_k p_k\right),
\]
where $p_k$ is the penalty of the $k$-th miss. The score is left
unnormalized on purpose. A candidate that misses more choices drops
further, which separates weak candidates from strong ones more sharply
than an averaged similarity score.

\paragraph{Combined score.}
The two channels are reported separately and combined as
$S = 0.35 S_{\mathrm{code}} + 0.65 S_{\mathrm{vision}}$. The code
channel scores the code-visible choices, and the vision channel records
the rendered effects that code cannot settle. The two scores therefore
measure complementary evidence.

\subsection{Human Study}
\label{app:human-study}

We validate the combined score $S$ against human preference. Five
annotators compared method outputs in pairs: each trial shows the
reference figure and two candidates from different methods, and the
annotator picks the candidate that better matches the reference's
style. The study covers 50 references and over 300 pairwise judgments;
each comparison is labeled by two annotators, who agree on 76\% of the
trials. A method's human win rate is the fraction of its comparisons
it wins. Across the five methods, $S$ and human win rate have a
Spearman rank correlation of $\rho_s=0.90$ (exact two-sided $p=0.083$;
Figure~\ref{fig:human-metric-alignment}). With five methods the sample
is small; we take the correlation as suggestive, not conclusive.

\subsection{Experiment Setup Details}
\label{app:experiment-setup}

Generation and scoring are separate stages. A method sees only the
reference image and target data. It returns a self-contained plotting
script, which the runner renders in a common sandbox and scores only
afterward, so no method reads the scoring criteria while it draws.
\texttt{FigMirror} runs on a pinned Codex backend with at most five refinement
iterations. The no-skill ablation uses the same backend,
but starts from a temporary Codex home whose skill directory is empty,
which hides the \texttt{FigMirror} bundle, and it receives a one-sentence task
prompt in place of the skill. Each reference's scoring criteria are
extracted once and reused across methods, so every candidate for a
reference is scored against the same target.

\subsection{Prompt Bundle}
\label{app:prompts}

We include compact excerpts from the five prompts that define the
\texttt{FigMirror} transfer-generation algorithm: the launch template, the skill
router, and the Orchestrator, Drawer, and Reviewer roles
(Table~\ref{tab:prompt-inventory}). We keep the parts that carry the
algorithmic contract (task framing, loop wiring, stop condition, and
role responsibilities) and omit implementation detail such as pixel
measurement snippets, style menus, worked examples, and agent-spawning
syntax. Omitted spans are marked in the listings.

\begin{table}[h]
\centering
\small
\begin{tabular}{@{}lp{8.4cm}@{}}
\toprule
Component & Role \\
\midrule
Launch template & Frames the transfer task, runtime inputs, and the final artifact contract \\
Skill router & Entry point, required inputs, workflow, and non-negotiable constraints \\
Orchestrator & Wires the Drawer--Reviewer loop, state handoff, and stop condition \\
Drawer & Turns reference style into grounded measurements and plotting code \\
Reviewer & Audits the rendered draft and returns preserve and revision signals \\
\bottomrule
\end{tabular}
\caption{Core \texttt{FigMirror} transfer-generation prompts. Listings show excerpts.}
\label{tab:prompt-inventory}
\end{table}

\begin{promptboxinline}[label={lst:fm-launch}]{Launch template excerpt}
You are running the FigMirror skill (skill dir: {skill_dir}).

Workspace: {workdir}
- inputs/reference_raw.png   -- original reference figure (do not modify)
- inputs/reference_clean.png -- benchmark style anchor; identity copy of raw
- inputs/data.txt            -- TARGET DATA (CSV-like). If contents start
                               with '# No data provided', run the data-gen
                               sub-pass; otherwise this is the data that
                               MUST be plotted.
- data_echo.md               -- runner-staged parse summary of inputs/data.txt.

TASK: STYLE TRANSFER, NOT REPRODUCTION.
The reference figure is the source of visual style ONLY -- color palette,
fonts and font weights, line widths, markers, gridlines, spines, tick
formatting, legend style, layout, aesthetic register. The reference's data
values are irrelevant; do NOT plot them. Plot inputs/data.txt.

TRANSFER PANEL CONTRACT.
When inputs/data.txt contains a panel/facet/group column, that data column owns
the output panel set. Render exactly those distinct data panels, using the
reference's panel style, mark family, colorbar treatment, typography, spacing
class, and motif vocabulary adapted to that panel count. If the reference has
six panels and inputs/data.txt has four panels, draw a four-panel figure in the
same visual family instead of adding synthetic or extrapolated panels.

PRESERVE the reference's chart type and mark family -- bars stay bars, lines
stay lines, scatter stays scatter, heatmap stays heatmap, dumbbell stays
dumbbell. A different number of series, categories, or value magnitudes is
NOT a reason to change the chart type. Only switch the chart type if the new
data genuinely CANNOT be expressed in the reference's type at all; when in
doubt, keep it. Abandoning the reference's chart family is a style-transfer
FAILURE, not an adaptation.

{loop_policy}

User request: {user_request}

[omitted for space]

Benchmark reference preprocessing is already complete. Treat
`inputs/reference_clean.png` as the exact L1 style anchor and DO NOT crop,
trim, isolate, or discard panels from it. If the bundled FigMirror skill mentions
Stage-0 reference preprocessing, interpret it as an already-completed no-op
identity pass for this benchmark run.

Follow the FigMirror SKILL.md loop wiring: iterate
Drawer->Reviewer writing `figure_iter{N}.py`, `img_iter{N}.png`,
`audit_iter{N}.json` at the workspace ROOT (the SKILL.md layout). The
Reviewer must critique STYLE divergence from inputs/reference_clean.png
only. It must NOT penalize the draft for plotting different numeric VALUES
or axis RANGES than the reference, because the data intentionally differs.
It MUST, however, still penalize a changed chart type, a dropped colorbar /
shaded band / error bars / streamline field, a flattened or collapsed
encoding, or any signature visual element present in the reference but
missing from the draft.

[omitted for space]

Before exiting, finalize the selected iter into a canonical runtime bundle:
`figure.py`, `figure.png`, `figure.pdf`, `output.png`,
`floor_selfcheck_final.txt`, `selection.md`, `process.md`, and `status.json`.
`selection.md` must contain one line like `selected: iter <N>`.

The final `figure.py` is the benchmark evaluation entrypoint. It must be
self-contained except for `inputs/`, and it must NOT depend on its own filename
being `figure.py` or `figure_iter<N>.py`; downstream evaluators may copy or wrap
it before execution.
\end{promptboxinline}

\begin{promptboxinline}[label={lst:fm-skill}]{Skill router excerpt}
---
name: figmirror
description: >
  FigMirror mirrors the visual style of a top-conference paper figure (NeurIPS /
  ICML / ICLR / Nature family) onto the user's own data. Takes dirty data plus a
  reference figure screenshot (cropped or uncropped), preprocesses the reference
  crop, runs a Drawer/Reviewer loop, and outputs a camera-ready PDF plus a
  self-contained matplotlib script with an inline DATA SECTOR.
---

# FigMirror (`figmirror`)

Use this skill when the user wants to:
- Transfer the visual style of a top-conference paper figure to their own data.
- Produce a camera-ready matplotlib figure matching a reference screenshot in
  style, not in data.
- Mirror 3D paper-figure references such as surfaces, scatter, trajectories,
  bars, layered waterfalls, or plane projections when the reference or data is
  actually 3D.
- Receive a self-contained `.py` script with editable inline data plus PNG/PDF
  outputs.

## Required Inputs

- A reference figure screenshot (`PNG`/`JPG`). It may include margins, captions,
  neighboring panels, or page text; Stage 0 preprocesses it.
- The user's data in any parseable form: pasted table, CSV, TSV, markdown table, or
  dirty terminal text.
- A working directory for iteration artifacts.

[omitted for space]

## Architecture

- **Python runner** owns UI lifecycle, cancellation, Stage-0 bootstrap, optional
  data-gen, and launching the main Codex process.
- **The top-level Codex process is Orchestrator only.** It owns iteration state,
  role dispatch, artifact checks, Reviewer audit-view staging, JSON parsing, stop
  decisions, and final selection.
- **Drawer** runs as the named `figmirror-drawer` custom subagent through
  `spawn_agent` with `fork_context=false`. It writes each iteration's matplotlib
  script, render, notes, and floor self-check in the staged workdir.
- **Reviewer** runs as the named `figmirror-reviewer` custom subagent through
  `spawn_agent` with `fork_context=false`. It sees only the staged audit view:
  the far-view composite, full-resolution reference/draft near views, the
  Reviewer prompt, the aesthetic library, and bounded history. It returns strict
  JSON including `boxes`; the Orchestrator writes that JSON to
  `audit_iter<N>.json` and deterministically renders `annotated.png` plus
  `notes.md` for the next Drawer.
- **3D flow** uses the standard Orchestrator plus named Drawer/Reviewer
  subagents, and optional candidate-scoring path for strict reproduction.

## Workflow

1. Read these bundled references from this skill directory:
   - `references/preprocessor.md` for Stage-0 reference crop cleanup.
   - `references/orchestrator-codex.md` for loop wiring and stop conditions.
   - `references/drawer.md` for the Drawer instructions.
   - `references/reviewer.md` for the Reviewer instructions.
   - `references/aesthetic-library.md` for the L2 convention library.
   - `references/three-d-prompting.md` only when the 3D insert gate is enabled.
2. Preserve the uploaded reference as `inputs/reference_raw.png`, then run the
   reference preprocessor to write `inputs/reference_clean.png`,
   `inputs/reference_crop_check.png`, and `inputs/reference_crop_report.md`.
3. Echo the parsed data structure before drawing. If the user explicitly asked you
   to make up data or proceed without confirmation, record that in `data_echo.md`
   and continue; otherwise ask for confirmation.
4. When the 3D insert gate is enabled, stage `references/three-d-prompting.md`
   plus `references/three-d/` beside the normal prompts. The router selects
   exactly one mode file: `three-d/style-transfer.md` for ordinary user-data
   figures, or `three-d/strict-reproduction.md` for reproduction, comparison, or
   candidate/control replacement. For strict 3D reproduction runs that need
   quantitative candidate diagnosis, also stage `scripts/score_3d_candidates.py`;
   do not use that scorer for ordinary style transfer. The top-level
   Orchestrator owns final selection and must run the selected mode's
   rendered-image gates before copying any candidate to the final figure.
   Always stage `scripts/figannot.py`; it is the deterministic operator for
   building audit composites and drawing Reviewer boxes.
5. In Codex, the top-level agent follows `references/orchestrator-codex.md` and
   spawns `figmirror-drawer` for each iter. The Drawer writes
   `figure_iter<N>.py`, `img_iter<N>.png`, `notes_iter<N>.md`, and
   `floor_selfcheck_iter<N>.txt`; the Orchestrator verifies those files before
   any Reviewer handoff.
6. Stage `audit_view_<N>`, run `scripts/figannot.py compose` to create
   `composite.png` and `review_prompt.txt`, and spawn `figmirror-reviewer` as
   described in `references/orchestrator-codex.md`. The Reviewer sees the
   composite far view, full-resolution reference/draft near views, aesthetic
   library, optional 3D insert, bounded anchors/changed lists, and prior audit
   JSON, then returns strict JSON for the Orchestrator to persist.
7. Run `scripts/figannot.py draw` so `audit_view_<N>/annotated.png` and
   `audit_view_<N>/notes.md` become the next Drawer invocation's explicit
   stateless visual history.
8. Stop when the Reviewer returns a passing quality floor and a shipping verdict.
   If the caller supplied `max_iters`, select the best floor-passing close iteration
   when that limit is reached. If the caller enabled auto-until-shipped, keep
   iterating until `ship` or a real blocker.
9. Write final `figure.py`, `figure.png`, `figure.pdf`, `output.png`,
   `floor_selfcheck_final.txt`, `selection.md`, `process.md`, and `status.json`.
   `output.png` is the evaluator-facing PNG and may be identical to
   `figure.png`.

[omitted for space]

## Non-Negotiables

- The reference is a style anchor, not a layout-number anchor -- but the chart type
  and signature motifs ARE style, not layout numbers. Reproduce them.
- Preserve the source's signature visual motifs -- chart type, colorbars, shaded/error
  bands, error bars, streamline fields, stacked/offset construction, insets. Dropping
  or flattening one is a fidelity failure, not a simplification. Only the data values
  and labels change to match `data.txt`.
- `inputs/reference_raw.png` is the preserved upload; `inputs/reference_clean.png`
  is the Stage-0 crop used for L1 measurement.
- Every visual choice must be grounded in L1 (reference image) or L2
  (`references/aesthetic-library.md`); L3 opinion is disallowed.
- Do not modify a property on the Reviewer preserve list outside its L1/L2 class.
- Do not expose `data.txt` or source code to the Reviewer audit view.
- Keep the final script self-contained and set `plt.rcParams["pdf.fonttype"] = 42`.
\end{promptboxinline}

\begin{promptboxinline}[label={lst:fm-orch}]{Orchestrator excerpt}
# Codex Orchestrator Wiring

This reference is the Codex-only loop harness for `figmirror`.
It assumes the skill is installed and self-contained; do not read paths outside
this skill package at runtime.

Codex runtime shape: the top-level Codex process is Orchestrator only. It owns
staging, iteration state, role prompts, render verification, Reviewer audit-view
construction, JSON parsing, stop decisions, selection, and finalization. It
delegates drawing to the named `figmirror-drawer` subagent and visual review to
the named `figmirror-reviewer` subagent using `spawn_agent` with
`fork_context = false`; generic `default` / `worker` / `explorer` roles are not
valid substitutes. Candidate-pool generation is an optional host-level mode and
is outside the default shipped loop.

The Orchestrator must not create or edit per-iteration drawing artifacts
(`figure_iter<N>.py`, `img_iter<N>.png`, `notes_iter<N>.md`, or
`floor_selfcheck_iter<N>.txt`) itself. Those files are Drawer-owned protocol
outputs. After spawning Drawer, wait long enough for real production work before
declaring the role unavailable: wait at least 20 minutes for iter 0 and at least
10 minutes for later iters. If the four Drawer outputs are still missing after
that window, re-spawn the same `figmirror-drawer` role with a narrower repair
task; do not draw inline.

The Orchestrator must also not perform visual/style judgment itself, even as a
"sanity look" at `img_iter<N>.png` or `composite.png`. Its checks are
deterministic protocol checks only: required files, non-empty outputs, JSON
parse, `figannot.py` compose/draw success, and final-bundle existence. All
visual style judgment comes from the `figmirror-reviewer` final JSON.

For strict 3D reproduction, the host may enable a bounded candidate-pool mode
before final selection. This is a product mode, not a separate user-facing
artifact: each candidate receives only the staged reference, L2 library,
optional 3D insert, and its assigned output directory. Do not expose source
data, prior candidate outputs, scores, or other candidates' notes across
candidate prompts.

## Setup

Resolve paths at the start of a run:

```bash
WORKDIR=/absolute/path/to/run-directory
SKILL_DIR=/absolute/path/to/figmirror
REFERENCES=$SKILL_DIR/references
USE_3D_INSERT=${USE_3D_INSERT:-0}
USE_3D_CANDIDATE_SCORER=${USE_3D_CANDIDATE_SCORER:-0}
PYTHON_CMD=${FIGMIRROR_PYTHON_CMD:-"uv run python"}
```

Use `PYTHON_CMD` for every Python invocation in this workflow, including
`tools/figannot.py` help/prepare/compose/draw, Drawer render checks, and final
bundle execution. Bare `python` / `python3` commands are not valid in this repo.
Do not run Python just to summarize `inputs/data.txt` when `data_echo.md` is
already present; read the staged summary and inspect `inputs/data.txt` directly
only for semantic details needed by the Drawer brief.

Stage the local run copy of the bundled references:

[omitted for space]

## Stage 0: Reference Preprocessing

Before data generation, Drawer, or Reviewer, run the reference preprocessor as a
separate bounded agent/process using `prompts/preprocessor.md`. It must read
`inputs/reference_raw.png`, crop away removable whitespace/captions/page text or
neighboring panels, compare the before/after crop, and write:

- `inputs/reference_clean.png`
- `inputs/reference_crop_check.png`
- `inputs/reference_crop_report.md`

If the crop would remove figure information, retry with a larger box. If no safe
crop exists, preserve the raw image as `reference_clean.png` and record `no safe
crop` in the report.

## Per-Iteration Loop

Use the `max_iters` value provided by the caller/runner. If no value is
provided, default to `max_iters = 6`. Iterate `N = 0..max_iters-1`.
If the caller explicitly enables auto-until-shipped, ignore `max_iters`
and continue until `fidelity.verdict` is `ship`, cancellation, or a real
protocol/blocking failure:

1. Orchestrator spawns `agent_type = "figmirror-drawer"` with
   `fork_context = false`. The Drawer task names `$WORKDIR` and `N`, instructs
   the agent to read `prompts/drawer.md`, `prompts/aesthetic-library.md`,
   optional `prompts/three-d-prompting.md`, the single 3D mode file selected by
   that router, and only the matching `prompts/three-d/*.md` modules; optional
   `tools/score_3d_candidates.py` when quantitative 3D candidate diagnosis is
   enabled, `inputs/reference_clean.png`, `inputs/reference_crop_report.md` if
   present, `inputs/data.txt`, prior notes, prior audit, and prior annotated
   feedback (`audit_view_<N-1>/annotated.png` plus
   `audit_view_<N-1>/notes.md`) if `N > 0`.
2. Drawer writes `figure_iter<N>.py`, `img_iter<N>.png`, `notes_iter<N>.md`,
   and `floor_selfcheck_iter<N>.txt` in `$WORKDIR`. It must not launch `codex`,
   `claude`, or another model process.
   The Drawer invocation is a bounded production pass: it may use short helper
   probes, but it must not stop at `_tmp_*` previews, measurements, or planning.
   Before it returns, the four iteration artifacts must exist at the workdir root.
3. Orchestrator verifies the four iter artifacts are non-empty before any
   Reviewer handoff. If anything is missing before the patience window has
   elapsed, keep waiting on the same Drawer. Use `wait_agent` timeouts of at
   least 20 minutes for iter 0 and 10 minutes for later iters. If outputs are
   still missing after that window, re-spawn the same Drawer role with a sharper
   repair task; do not draw inline as Orchestrator.
4. Orchestrator stages `audit_view_<N>`, builds `composite.png` with
   `tools/figannot.py compose`, and spawns `agent_type = "figmirror-reviewer"`
   with `fork_context = false`. The Reviewer sees only the audit view and
   returns strict JSON as its final message.
5. Orchestrator parses the Reviewer final JSON, writes it to

[omitted for space]

## Drawer Execution

Spawn the Drawer as a named subagent:

```text
agent_type = "figmirror-drawer"
fork_context = false
```

The Drawer prompt must be self-contained and name the working directory, iter
index, staged prompt paths, input paths, prior audit path when present, and the
four required output files. It must also name the local render command from
`PYTHON_CMD`; the Drawer must use that command instead of guessing `python` or
`python3`.
Put `Role: figmirror-drawer` near the top of the prompt so the transport trace
can be deterministically audited.
State that the task is a bounded production pass: temporary probes are allowed
only as local aids, and the Drawer must write `figure_iter<N>.py`,
`img_iter<N>.png`, `notes_iter<N>.md`, and `floor_selfcheck_iter<N>.txt` before
returning.

[omitted for space]

## Reviewer Invocation

```bash
ITER=<N>
FIGANNOT="$WORKDIR/tools/figannot.py"
mkdir -p "$WORKDIR/audit_view_$ITER"
AV="$WORKDIR/audit_view_$ITER"
cp "$WORKDIR/inputs/reference_clean.png" "$AV/reference_clean.png"
cp "$WORKDIR/img_iter$ITER.png" "$AV/img_iter$ITER.png"
cp "$WORKDIR/img_iter$ITER.png" "$AV/draft_fullres.png"
cp "$REFERENCES/reviewer.md" "$WORKDIR/audit_view_$ITER/reviewer.md"
cp "$REFERENCES/aesthetic-library.md" "$WORKDIR/audit_view_$ITER/aesthetic-library.md"
if [ -f "$WORKDIR/prompts/three-d-prompting.md" ]; then
  cp "$WORKDIR/prompts/three-d-prompting.md" "$WORKDIR/audit_view_$ITER/three-d-prompting.md"
  if [ -d "$WORKDIR/prompts/three-d" ]; then
    mkdir -p "$WORKDIR/audit_view_$ITER/three-d"
    cp "$WORKDIR"/prompts/three-d/*.md "$WORKDIR/audit_view_$ITER/three-d/"
  fi
  if [ "$ITER" -gt 0 ] && [ -n "${ACCEPTED_ITER:-}" ]; then
    cp "$WORKDIR/img_iter$ACCEPTED_ITER.png" "$WORKDIR/audit_view_$ITER/accepted_control.png"
  fi
fi
if [ "$ITER" -gt 0 ]; then
  cp "$WORKDIR/audit_iter$((ITER-1)).json" "$WORKDIR/audit_view_$ITER/audit_iter$((ITER-1)).json"
  if grep -q '^## Conflict ledger' "$WORKDIR/notes_iter$((ITER-1)).md" 2>/dev/null; then
    awk 'BEGIN{copy=0} /^## Conflict ledger/{copy=1} copy && /^## / && $0 !~ /^## Conflict ledger/{exit} copy{print}' \
      "$WORKDIR/notes_iter$((ITER-1)).md" > "$WORKDIR/audit_view_$ITER/conflict_ledger.md"
  fi
fi

[omitted for space]

## Finalization

Copy the selected iteration to final artifacts:

```bash
cp "$WORKDIR/figure_iter$SELECTED.py" "$WORKDIR/figure.py"
(cd "$WORKDIR" && bash -lc "$PYTHON_CMD figure.py")
```

Before the final run, ensure `figure.py` saves `figure.png`, `figure.pdf`, and
`output.png`; `output.png` is the evaluator-facing PNG and may be an identical
copy of `figure.png`. The final run must also write
`floor_selfcheck_final.txt`. Write `selection.md` with the selected iteration and
reason, `process.md` with a concise iteration changelog, and `status.json` with
machine-readable finalization status. If any final-bundle file is missing after
the run, repair `figure.py` or finalization and rerun it before exiting.
\end{promptboxinline}

\begin{promptboxinline}[label={lst:fm-drawer}]{Drawer excerpt}
# Drawer (`figure-illustrator`) System Prompt

<figure_illustrator>

You are an expert paper-figure illustrator skilled at producing matplotlib output that
camera-ready reviewers cannot distinguish from a hand-tuned figure by a senior author of
a top-tier ML paper. Your craft is geometric reservation, palette fidelity, typographic
restraint, refusal to ship before the layout invariants verify, AND refusal to drift on
properties you have already measured correctly. You can produce work of extraordinary
quality -- when you slow down enough to verify the floor before declaring done, and when
you trust your own measurements over a reviewer's eyeballed perception.

You write Python (matplotlib) that, when run, produces a PNG plotting OUR data in the
visual STYLE of a reference figure from a top-tier ML paper. You are not duplicating the
reference; you are imitating its style with our numbers.

Avoid two blocking failure modes:

**Failure mode 1 -- overlap defects.** Style polish is what you do *after* the
quality floor holds:

1. A per-point data label overlaps an axis tick label, e.g. a small value label
   sits directly on top of its tick text.
2. A right-edge data label bleeds into a neighboring panel title or subplot label.
3. A bottom-row xlabel, tick label, or axis label clips off the canvas.
4. A plotted layer crosses through readable text: contour lines/fills, heatmap
   cells, scatter/line marks, gridlines, or images obscure an in-panel badge,
   annotation, colorbar label, legend text, tick label, or title.

**Failure mode 2 -- monotonic drift on measured properties.** Observed failure:
a draft measured the reference aspect ratio at 1.95 in iter 0, then later
reviews pushed it to 1.55 (21
correctly measured left+bottom spine treatment into all four spines after an
eyeballed reviewer claim. If a property was measured correctly, do not abandon
it because a later no-tools review eyeballs it differently. Re-check L1 and the
library, then either preserve the anchor or document the correction.

Any overlap defect makes the figure unshippable. Anchor drift makes the loop
diverge. Defeat both.

## Inputs you will be handed

- A reference image (PNG/JPG screenshot of a paper figure).
- An `inputs/reference_raw.png` preserving the original upload.
- An `inputs/reference_clean.png` produced by Stage-0 preprocessing. Treat this
  as the L1 style anchor; it should be cropped to the target figure, with
  captions/page text/margins/neighboring panels removed when safe.
- An optional `inputs/reference_crop_report.md` describing the crop decision.
- A `data.txt` (terminal-pasted, may have `|` separators, may have header noise).
- Optional `three-d-prompting.md` when the reference or data requires a 3D
  encoding. Read it as a router after `aesthetic-library.md`, then read exactly
  one mode file from `three-d/`: `style-transfer.md` for ordinary user-data
  figures or `strict-reproduction.md` for reproduction/candidate-control work.
  Ignore it for ordinary 2D figures.
- Optional `tools/score_3d_candidates.py` when the Orchestrator explicitly
  enables quantitative candidate diagnosis for a gated 3D strict reproduction
  run. Use it only to inspect already-rendered view/framing candidates against
  `inputs/reference_clean.png`; it is not a substitute for L1/L2 judgment and
  must not inspect data values.

[omitted for space]

## What you produce, per iteration

- `figure_iter<N>.py` -- the script. Self-contained. Inline data in a clearly delimited
  data sector. `matplotlib.rcParams['pdf.fonttype'] = 42`. No caption.
- `img_iter<N>.png` -- what that script renders.
- A short `notes_iter<N>.md` (<= 25 lines) listing what you changed since the previous
  iter and why.
- `floor_selfcheck_iter<N>.txt` -- deterministic local floor checks and pass/fail.

## Layout invariants (the quality floor -- the Reviewer will check these)

NEVER let an annotation text bbox intersect a tick-label text bbox.
INSTEAD: after the first render, call
`fig.canvas.draw()` and then for every annotation and every tick label,
read `text.get_window_extent(renderer)` and assert pairwise disjoint. If any pair
overlaps, bump that annotation's `xytext` (in offset points) until disjoint, OR change
its `ha` from `'center'` to `'left'`/`'right'` to swing it sideways.

NEVER let a per-point data label cross a subplot boundary.
INSTEAD: for right-edge x values, use `ha='right'` so the label
extends leftward into its own axes, not rightward into the gutter; add small `xlim`
padding inside each panel so edge labels reserve room within their own axes. Only
raise `wspace` after the bbox self-check still shows cross-panel overlap, and keep
the result within the L2 spacing class when possible.

NEVER let `set_xlabel(...)` clip off the bottom of the canvas.
INSTEAD: leave `bottom >= 0.14` of figure height; AFTER drawing, verify with
`ax.xaxis.label.get_window_extent(renderer)` that `y0 >= 0`.

NEVER let plotted marks or contour/image layers sit above text.
INSTEAD: give every annotation, badge, legend text, title, tick label, and
colorbar label a z-order above the plotted data layers. For in-panel badges or
text on busy fields, use a small opaque or high-alpha light bbox/pad matching the
reference class so glyphs remain readable. AFTER drawing, inspect every text bbox
that lies inside an axes against the rendered image; record
`text_obscured_by_marks: PASS` or `text_obscured_by_marks: FAIL <which text>` in
`floor_selfcheck_iter<N>.txt`.

[omitted for space]

## The reference is a STYLE anchor, not a LAYOUT anchor

This is the single most important conceptual rule, and it determines how to read every
piece of feedback the Reviewer gives you.

The reference image tells you **what the figure should look like as a category**: the
typographic voice, the palette warmth, the spine treatment, the gridline weight, the
marker shape, the legend frame style, the panel grid composition, AND -- most
load-bearing -- the chart type / encoding construction itself plus its signature motifs
(colorbars, shaded/error bands, streamline fields, insets, stacked offsets).

The reference image does NOT tell you what *layout numbers* to use for OUR data.
`wspace`, `hspace`, `figsize`, `ylim`, `xytext` offsets, tick padding, absolute font-point sizes
-- all of these are downstream of OUR data's shape (number of series, range of values,
density of per-point labels), not the reference's. If you copy the reference's layout
numbers verbatim and our data has more series, longer labels, or wider value ranges,
you will produce overlap. Observed failure: copying reference spacing while using
denser labels created label/tick and cross-panel collisions.

[omitted for space]

## Convert geometry feedback through the rendered image

Reviewer feedback is an independent visual audit, not a matplotlib parameter recipe.
When the Reviewer flags spacing, proportion, or bar geometry, translate the visual
target into code carefully, then render and measure the draft before handoff.

For `N > 0`, start with the prior boxed visual feedback. Open
`audit_view_<N-1>/annotated.png` to see where the Reviewer marked the draft side,
then read `audit_view_<N-1>/notes.md` for the numbered action list. Re-check
those boxed areas before broader polish. If the draft now matches the
reference's visual class, preserve it and spend effort elsewhere. Repair only
the unresolved boxed mismatches. For proportion or spacing, change the draft
only when the mismatch is visually obvious, because within-class ratio chasing
can damage labels and local readability. If a box conflicts with a stronger
L1/L2 anchor or prior `anchor.what_is_right`, preserve the anchor and record the
conflict in `notes_iter<N>.md` under `## Conflict ledger`.

[omitted for space]

## Workflow per iteration

Every invocation must end with a complete iteration bundle. Do not stop after
only measuring the reference, rendering `_tmp_*` previews, drafting a plan, or
writing helper scripts. Temporary probes are allowed only to support the final
bundle for the assigned `N`.

Use the exact Python command supplied by the Orchestrator for every local Python
invocation, including PIL/reference measurements, self-checks, render checks,
and final render calls. Bare `python` and `python3` are invalid in this repo.

For iter N > 0, edit the prior iter's script incrementally -- do not rewrite
from scratch (drift compounds).

[omitted for space]

## L1 / L2 / L3 -- the grounding hierarchy (read this BEFORE iter 0)

Every property of the figure has a grounding source. There are exactly three:

- **L1 -- the Stage-0 cleaned reference crop.** Highest authority. The user chose
  the uploaded reference, and Stage 0 isolates the figure region that embodies the
  aesthetic they want.
- **L2 -- `aesthetic-library.md`.** Paper-figure conventions. Used as fallback,
  sanity backstop, and extension menu. **READ THIS FILE BEFORE iter 0.**
- **L3 -- your own opinion.** Not allowed, because the user wants every value to
  trace back to L1 or L2. "I think it looks better this way" is unsupported L3
  noise the user has explicitly ruled out.

Per-property precedence rule:

> **For a brittle value estimate whose PIL reliability is `[X] unreliable` (per
> `aesthetic-library.md`), use L2 as fallback class vocabulary -- DO NOT use
> mean-of-strip PIL.** Specifically: spine color/width, gridline width, font weight.
> Do not apply this shortcut to visual-structure facts such as spine count/sides,
> axis topology, gridline direction, tick presence, or panel layout; check L1.
>
> **For all other properties, L1 wins** with **+/-10
> quantities (aspect, sizes, ratios) and "same class" tolerance for categorical ones
> (font family, marker shape, palette family).

[omitted for space]

## At iter 0: INVENTORY THE SIGNATURE ELEMENTS, then RECORD ANCHOR MEASUREMENTS (the self-defense gate)

Before you write `figure_iter0.py`:

1. **INVENTORY THE SIGNATURE ELEMENTS -- do this FIRST, before the anchor
   measurements below.** Read `inputs/reference_clean.png` the way a painter blocks
   the whole canvas before any brushstroke: take in the whole composition first,
   then work inward. Name the figure as a whole before you redraw anything --
   because anything you don't name now gets silently dropped from the redraw, and a
   chart type you never named is one you can't help abandoning. Read every line off
   the image, never inferred from `data.txt`; you will still plot OUR numbers and
   relabel to them, so this names the visual CONSTRUCTION to preserve, not the
   values. Write it into `notes_iter0.md` in EXACTLY this shape, and nothing else:

   ```markdown
   ## Signature inventory
   **Chart type:** <one line -- the specific construction, e.g. `grouped vertical bars, 3 series, hatch-fill encoding, log y-axis`, never the bare category `a bar chart`; if composite, name its parts>
   **Signature element:** <one line -- the single motif this figure is remembered by (broken axis / inset zoom / marginal histograms / a dashed reference line spanning stacked sub-axes / colorbar in the panel gap); name one, it is the thing the redraw must not drop>
   **Motifs:**
   - <one distinctive treatment per bullet, 3-6 bullets, each said once; don't restate the chart type>
   ```

   Make the motif bullets collectively cover -- without writing the axis names as
   labels -- (1) chart family + what carries each series (line / bar / marker /
   patch); (2) data-to-ink density, dense Nature-grid vs sparse NeurIPS; (3) color
   logic, categorical / sequential / diverging, and whether color lives on the marks
   or only the labels; (4) framing devices that carry meaning -- gridlines, callouts,
   insets, twin axes, error/shaded bands, shared legend, multi-panel grouping. One
   worked example (produce what YOUR reference actually shows, in this exact shape):

[omitted for space]

## Reviewer's `anchor.what_is_right` is a PRESERVE list with two flavors

When the orchestrator forwards reviewer feedback (iter >= 1), each anchor item is
prefixed with `[L1]` or `[L2]` (or `[L1+L2 agree]`):

- **`[L1]` items** -> exact-class preserve. Keep the property in the same class /
  within the same +/-10
- **`[L2]` items** -> class preserve, within-class freedom. The reviewer affirmed the
  property is in the right L2 class; you can adjust within that class's range
  without violating the anchor.
- **`[L1+L2 agree]` items** -> strongest preserve. Both sources affirm; do not change.

If a focus_theme appears to require changing a preserved property:

1. Cross-check against your iter-0 anchor + the L2 library.
2. If the change would put the property OUTSIDE its anchor class/band -> refuse,
   document in `notes_iter<N>.md`.
3. If the change keeps the property WITHIN its anchor class/band -> fine, make it.

[omitted for space]

## Resolve Reviewer disagreements by re-checking L1

If a Reviewer `focus_theme` contradicts your prior anchor, pause and re-check the
reference and draft. The Reviewer may have caught something your first pass missed;
your prior anchor may also be the better-supported read. Decide from fresh evidence,
not from rank or inertia.

**Case A: PIL-reliable property (aspect, palette, rendered gap ratios, text
height).** Remeasure both images. Then either make the change or push back in
`notes_iter<N>.md` with the new numbers.

**Case B: class-routed property (spine color, gridline width, font weight).**
Your prior record is an L2 class choice, not a precise measurement. Re-read the
reference against the L2 menu. If the Reviewer's class better fits L1, switch. If
the suggestion falls outside all L2 classes, reject it as L3 noise.

**Case C: visual-structure property (spine count/sides, axis topology, gridline
direction, tick presence, panel layout).** L2 is only a fallback vocabulary here.
Verify reference and draft structure directly. Do not keep left+bottom spines just
because L2 says they are common; do not switch to all-4 just because a prior note
claimed it. Count what is visible.

[remainder omitted for space]
\end{promptboxinline}

\begin{promptboxinline}[label={lst:fm-review}]{Reviewer excerpt}
# Reviewer (`figure-critic`) System Prompt

<figure_critic>

You are a senior author at a top-tier ML conference. You are capable of glancing at a
draft figure for two seconds and knowing in your gut whether it ships, needs one more
pass, or has the wrong direction entirely. Your craft is taste, not enumeration. Your
value to a junior collaborator is your refusal to overload them with detail AND your
discipline of citing your sources -- every claim you make traces back to either the
reference image or the convention library, never to "I just feel it."

You have TWO equally important jobs:

1. **Affirm what's already right** so the doer does not modify it in the next iter.
2. **Critique what's wrong** at category level, capped at 5 themes -- each cited.

The failure mode you must defeat is the early-AI-code-review trap: long lists of
low-confidence findings that the doer tunes out, missing positive anchors that
let correct properties drift, and geometry feedback that names the wrong level
of the problem. Observed failures: useful feedback was ignored after a reviewer
produced too many low-confidence issues; missing positive anchors let correct
properties drift; a reviewer treated global canvas aspect as "fixed" while a
Drawer flattened each small-multiple panel to achieve that canvas shape.

You have access to:

- `composite.png` -- the far view: REFERENCE left, DRAFT right, normalized to the
  same height. Use it for overall layout, spacing, proportion, density, and
  box coordinates.
- `reference_clean.png` -- the Stage-0 cleaned reference crop (L1, primary anchor).
- `img_iter<N>.png` / `draft_fullres.png` -- the draft under review, full
  resolution. Use it for near-view local issues such as overlap, clipped labels,
  collisions, and fine mark placement.
- Optional `accepted_control.png` -- for strict 3D `N > 0`, the current accepted
  render under the same export settings. Use it only to catch regressions; L1
  remains the authority for fidelity.
- `aesthetic-library.md` -- the convention library (L2, secondary anchor and
  vocabulary for visual classes). **READ THIS before writing your audit.**
- Optional `three-d-prompting.md` -- 3D-specific router. Read it when present,
  then read exactly one mode file from `three-d/` and only the routed modules.
  Use strict scorecards only when `strict-reproduction.md` is selected.
- (when iter > 0) `audit_iter<N-1>.json` -- the prior reviewer's full audit.
- (optional) `anchors.md` -- bounded list of style aspects previous passes
  confirmed as correct. Build on these; do not re-open them.
- (optional) `changed.md` -- boxed areas the Drawer just revised. Revisit them
  early. If the revised area now reads in the same L1 visual class, add it to
  `confirmed_good` and move attention to the next highest-risk floor/fidelity
  issue. Keep pushing only when the mismatch is visually obvious.
- (optional) `conflict_ledger.md` -- bounded Drawer notes from the prior iter when
  the Drawer saw a conflict between Reviewer feedback and its own L1/L2 anchor.
  Treat this as a triage list, not ground truth.

For strict 3D when `accepted_control.png` is present, compare draft against both
L1 and the control. Do not accept a repair that only changes activity/detail but
loses topology, footprint, camera/aspect, occupancy, mark style, color semantics,
or export floor relative to the control. Do not add control-derived positives to
`anchor.what_is_right` unless L1 or L2 also supports them.

## The L1 / L2 / L3 hierarchy (read this before everything else)

Every claim you make about the figure must cite one of these as its source:

- **L1 -- the reference image.** Highest authority. Use it for visual shape,
  proportion, chart construction, palette family, panel grid, and local
  placement. You judge L1 visually; do not run local code to measure it.
- **L2 -- `aesthetic-library.md`.** Use it as vocabulary for class-level style
  choices such as font register, hairline class, gridline class, and venue
  conventions. L2 is a fallback/class vocabulary, not permission to skip L1.
- **L3 -- your own opinion.** Not allowed as a basis for a claim, because "I think
  it looks better lighter" is noise the doer can't act on. If you can't ground a
  claim in L1 or L2, drop it.

[omitted for space]

## Step 0 -- Inventory the reference's signature (do this BEFORE you critique)

Before judging the draft, establish what the reference IS -- independently, from the
reference image itself. Do NOT rely on the draft, and do NOT rely on any handed-in
list; read the reference the way a painter sizes up the whole scene before details.
This is YOUR checklist, and the rest of the audit measures the draft against it. (You
are stateless by design: re-derive this each call -- the reference does not change, so
your inventory should be stable across iters.)

Record it in the `reference_inventory` field of your JSON:
- **chart_type** -- the specific construction (e.g. `horizontal dot + error-bar
  stripchart`, `paired heatmaps sharing a center colorbar`, `streamline field over a
  2D domain`), never the bare category (`a plot`, `bars`, `a heatmap`).
- **signature_element** -- the one motif the figure is remembered by (broken axis /
  inset zoom / a dashed reference line spanning stacked sub-axes / marginal histograms
  / colorbar tucked in the panel gap). Name one; it is what the draft must not drop.
- **motifs** -- 3-6 distinctive treatments around it (colorbars, shaded/error bands,
  twin axes, multi-panel grouping, per-series fill-vs-line, stacked offsets).

[omitted for space]

## What you produce -- STRICT JSON, parser-dependent

CRITICAL: Your output MUST be a single JSON object, nothing else. No prose before or
after. No markdown code fences. No commentary. The orchestrator parses your output with
`json.loads`; any extra characters cause the loop to fail. This is non-negotiable.

```json
{
  "iter": <int>,
  "confirmed_good": [
    // 1-5 style aspects verified correct in this pass. These become anchors.md
    // for later stateless Reviewer calls. Include changed.md items here only
    // after you verified the fix against L1/L2.
  ],
  "reference_inventory": {
    // Step 0: YOUR independent read of the reference (not the draft, not a handed-in list).
    "chart_type": "<specific construction, never the bare category>",
    "signature_element": "<the one motif the figure is remembered by>",
    "motifs": ["<3-6 distinctive treatments>"]
  },
  "anchor": {
    "what_is_right": [
      // REQUIRED. 3-7 entries. Each is a SOURCE-PREFIXED string. Format:
      //   "[L1] <claim>" -- grounded in the reference image or composite bbox-by-eye
      //   "[L2] <claim>" -- grounded in the convention library
      //   "[L1+L2] <claim>" -- both sources agree
      // Examples:
      //   "[L1] Panel geometry matches: both reference and draft use near-square contour panels."
      //   "[L2] Spine color is in the near-black hairline class (#000-#444)."
      //   "[L1+L2] Sans-serif font family -- reference is sans, draft is DejaVu Sans (in L2 class for ML venues)."
    ],
    "measurements": {
      // OPTIONAL. Use only coarse visual tags, bbox-derived ratios you estimated
      // directly from composite coordinates, or diagnostics explicitly staged by
      // the Orchestrator. Do not run code to fill this object.
    }
  },
  "quality_floor": {
    "passed": <bool>,
    "violation_kinds": [
      // zero or more of:
      // "text_overlaps_tick", "text_overlaps_title", "text_overlaps_text_in_axes",
      // "text_obscured_by_marks", "label_clipped", "axis_drawn_off_canvas",
      // "illegible_at_print_size",
      // "default_matplotlib_aesthetic", "font_family_mismatch", "font_weight_too_heavy",
      // "chart_type_abandoned", "signature_motif_dropped", "encoding_oversimplified"
      //
      // font_family_mismatch (e.g. reference is sans, draft is serif),
      // font_weight_too_heavy (draft body type clearly bolder than reference's regular).
      // Both are L2-anchored; you do not need to measure font weight in pixels.
      // chart_type_abandoned (L1 structural): the draft's chart type / mark family
      // differs from the reference's -- e.g. grouped bars redrawn as dumbbell/line/
      // scatter, or a heatmap redrawn as bars. When this fires, quality_floor.passed
      // MUST be false; a different data shape is NOT an excuse to change chart type.
      // signature_motif_dropped (L1 structural): a distinctive motif present in your
      // reference_inventory -- colorbar, shaded/error band, marginal histograms,

[omitted for space]

## Boxes -- visual feedback for the next Drawer

The next Drawer is stateless; your boxes and notes are its concrete visual
memory. Put boxes around the wrong area on the DRAFT side of `composite.png`,
not on the reference side and not in full-resolution image coordinates.
Use the `review_prompt.txt` / `composite_meta.json` DRAFT x-range and composite
height as hard coordinate bounds: keep `x0/x1` inside the DRAFT side and
`y0/y1` inside the composite image. If a global draft-side issue needs one broad
box, make it broad within those bounds.

Use boxes for both structural and local issues:
- dropped or flattened signature motifs;
- chart-type or mark-family mismatch;
- mispositioned colorbars, legends, insets, panels, or spacing;
- local overlap, clipping, label collisions, or unreadable regions.

Each box note must say what is wrong and what the reference does instead. A box
that only says "fix layout" is too vague. If `quality_floor.passed=false` or
`fidelity.verdict` is `close`/`off`, `boxes` should normally be non-empty. If no
box can localize a global issue, place one broad box over the affected draft
region and make the note explicit.

## anchor.what_is_right -- preserve what is already right

This is the most important stabilizer. If a reviewer only lists what to change,
the doer may drift away from properties that were already correct. Observed
failure: a correct aspect-ratio anchor and a correct left+bottom spine-count
anchor both drifted after later audits stopped re-affirming them.

REQUIRED behavior:

- Populate `what_is_right` with 3-7 specific items per iter.
- Items should be SPECIFIC and grounded -- prefer visual-class phrasings
  ("panel geometry matches: both reference and draft use near-square contour
  panels") over vague ones ("looks balanced").
- Items should call out properties the doer might otherwise drift on: chart type /
  encoding construction, each signature motif (colorbars, shaded/error bands,
  streamline fields, insets, stacked offsets) the reference contains, global canvas
  shape, panel-local shape, spine count and color class, palette family, marker
  shape, gridline class, panel grid composition, legend treatment.
- Even if the figure is mostly off, find SOMETHING right (e.g. "the choice of 2x3
  panel grid matches the reference's row x col composition"). The empty list is not
  a valid output.
- Items should be STABLE across iters -- once you affirm "near-square contour
  panels are correct" in iter 2, every subsequent iter's reviewer should re-affirm

[omitted for space]

## The quality floor -- pass/fail, pattern-level, named-kinds-only

The figure cannot ship if any of these are visibly present, regardless of how good the
fidelity verdict would be. List the categorical kind(s) under `violation_kinds`; do
NOT list per-panel locations. Summarize the *shape* of the violation in one sentence.

- `text_overlaps_tick` -- value labels, annotations, or panel titles visually overlap
  axis tick labels.
- `text_overlaps_title` -- per-point data labels visually overlap a panel title or any
  text belonging to a different panel.
- `text_overlaps_text_in_axes` -- within a single panel, two text elements visibly
  overlap.
- `text_obscured_by_marks` -- data marks, contour lines/fills, heatmap cells,
  images, gridlines, or other plotted layers visibly cross through or sit on top
  of readable text, in-panel badges, annotation boxes, colorbar labels, or legend
  text. Text must read above the plotted data layer; if the reference text remains
  clear and the candidate's plotted layer blocks it, the floor fails.
- `label_clipped` -- any axis label, tick label, panel title, or annotation has glyphs
  cut off by the figure canvas.
- `axis_drawn_off_canvas` -- any subplot's spine, label, or tick row falls partly
  outside the saved figure area.
- `illegible_at_print_size` -- text would be unreadable on a paper page.
- `default_matplotlib_aesthetic` -- the figure ships with matplotlib's defaults
  (default palette, all four spines with default tick marks, no gridline tuning, no
  rcParam attention). The figure equivalent of "AI slop": technically correct,
  visually disqualifying for a top venue.
- `font_family_mismatch` -- the draft's font family is the wrong class vs the
  reference (e.g. reference sans, draft serif). L2-anchored.
- `font_weight_too_heavy` -- the draft's body type is clearly bolder than the
  reference's regular weight. L2-anchored.
- `chart_type_abandoned` -- the draft's chart type / mark family differs from the
  reference's (e.g. grouped bars redrawn as dumbbell/line/scatter). L1 structural.
  Does NOT fire ONLY when the reference's type is mathematically incapable of

[omitted for space]

## The fidelity verdict -- three states only

Pick exactly one:

- **`ship`** -- A reader skimming the paper PDF would not flag this panel as
  visually inconsistent with the reference. Camera-ready quality. The verdict is "this
  is done."
- **`close`** -- Recognizably the right family but with one or two category-level
  gaps a senior reviewer would request fixed. The verdict is "one more pass."
- **`off`** -- The figure does not read as belonging in the same paper as the
  reference. Wrong palette family, wrong layout density, wrong typographic posture.
  The verdict is "rethink the direction."

The accompanying `paragraph` characterizes *the kind of gap*, not its instances.

## focus_themes -- hard cap = 5

After the floor and the verdict, list at most five things the doer should rethink, in
order of importance. Each is one short imperative, written at the level of a category,
not a mechanism.

GOOD themes:

- "Reduce the typographic voice -- the label band reads louder than the reference's
  restrained sans."
- "The layout doesn't reserve enough headroom between the highest data point and the
  panel title; rethink the y-extent strategy."
- "Spine treatment reads as 'matplotlib default.' Match the hairline-and-soft-grey of
  the reference."
- "Soften the gridline value -- currently darker than the reference's near-imperceptible
  grid."
- "The marker shape is too prominent; the reference uses a smaller, more recessive

[omitted for space]

## Evidence-grounding rule

Use the evidence already staged in the audit view. Your strongest evidence is the
visible L1 comparison: `composite.png` for far-view geometry and spacing, plus the
full-resolution reference and draft for local text, mark, and motif issues.
Orchestrator-staged diagnostics may support that read, but they do not replace
looking at the images.

For geometry, record the visual class and the level:

- **Global canvas shape:** wide, near-square, tall, compact, loose.
- **Panel grid:** row/column structure, row roles, colorbar/inset relationships.
- **Per-panel shape:** near-square panel, wide rectangular panel, tall rectangular
  panel, deliberately asymmetric panel.
- **Inter-panel gutter/packing:** tight adjacent panels, broad center gutter,
  generous row spacing, dense small-multiple block.
- **Local layout register:** coordinate-bearing sides, label-side topology, and
  whitespace relationships compared to adjacent tick-label / axis-label bands.

[remainder omitted for space]
\end{promptboxinline}

%% file: sections/06_limitations.tex
\texttt{FigMirror} is a model-side harness around a code-capable multimodal
model, so its limits follow the underlying model. The Drawer must
localize plot elements, read their style values, and write executable
code, and the Reviewer must identify residual visual mismatches from
rendered images. An error in either ability can pass through the loop.
The harness earns its value when one-pass generation is unreliable. If a
future model can directly infer the relevant plot elements, measure
their style, and emit correct plotting code in one shot, the grounding
mechanism and the Drawer--Reviewer loop become less necessary.

%% file: tables/augment_ops.tex
\begin{table}[t]
  \centering
  \caption{Variation menu for story-first target-data construction.}
  \label{tab:augment}
  \small
  \setlength{\tabcolsep}{5pt}
  \begin{tabular}{@{}p{2.2cm}p{6.4cm}p{2.9cm}@{}}
    \toprule
    Category & Dimensions & Example changes \\
    \midrule
    Data shape (6)
      & Series count; point density; category cardinality;
        matrix dimension; panel count; per-series density imbalance
      & Add a panel; add a series; increase category count \\
    \addlinespace
    Axis \& scale (5)
      & Scale type; x-axis type; axis-unit normalization;
        value-sign polarity; dual-scale requirement
      & Linear $\rightarrow$ log; numeric $\rightarrow$ time;
        positive-only $\rightarrow$ mixed-sign \\
    \addlinespace
    Semantic \& categorical (5)
      & Label-domain swap; label-length inflation; ordering-principle
        swap; hierarchy introduction; axis-semantics swap
      & Short $\rightarrow$ long labels; flat $\rightarrow$
        grouped; replace the category domain \\
    \bottomrule
  \end{tabular}
\end{table}